# HIGH-DIMENSIONAL ADDITIVE MODELING

By Lukas Meier, Sara van de Geer and Peter Bühlmann

*ETH Zürich*

We propose a new sparsity-smoothness penalty for high-dimensional generalized additive models. The combination of sparsity and smoothness is crucial for mathematical theory as well as performance for finite-sample data. We present a computationally efficient algorithm, with provable numerical convergence properties, for optimizing the penalized likelihood. Furthermore, we provide oracle results which yield asymptotic optimality of our estimator for high dimensional but sparse additive models. Finally, an adaptive version of our sparsity-smoothness penalized approach yields large additional performance gains.

**1. Introduction.** Substantial progress has been achieved over the last years in estimating high-dimensional linear or generalized linear models where the number of covariates $p$ is much larger than sample size $n$. The theoretical properties of penalization approaches like the lasso [28] are now well understood [3, 14, 23, 24, 33] and this knowledge has led to several extensions or alternative approaches like adaptive lasso [34], relaxed lasso [22], sure independence screening [12] and graphical model based methods [6]. Moreover, with the fast growing amount of high-dimensional data in, for example, biology, imaging or astronomy, these methods have shown their success in a variety of practical problems. However, in many situations, the conditional expectation of the response given the covariates may not be linear. While the most important effects may still be detected by a linear model, substantial improvements are sometimes possible by using a more flexible class of models. Recently, some progress has been made regarding high-dimensional additive model selection [7, 19, 26] and some theoretical results are available [26]. Other approaches are based on wavelets [27] or can adapt to the unknown smoothness of the underlying functions [2].

In this paper, we consider the problem of estimating a high-dimensional generalized additive model where $p \gg n$. An approach for high-dimensional additive modeling is described and analyzed in [26]. We use an approach which penalizes both the sparsity and the roughness. This is particularly important if a large number of basis functions is used for modeling the additive components. This is similar to [26] where the smoothness and the sparsity is controlled in the backfitting step. In addition, our computational algorithm, which builds upon the idea of a group lasso problem, has rigorous convergence properties and thus, it is provably correct for finding the optimum of a penalized likelihood function. Moreover, we provide oracle results which establish asymptotic optimality of the procedure.

**2. Penalized maximum likelihood for additive models.** We consider high-dimensional additive regression models with a continuous response $Y \in \mathbb{R}^n$ and $p$ covariates $x^{(1)}, \ldots, x^{(p)} \in \mathbb{R}^n$ connected through the model

$$Y_i = c + \sum_{j=1}^{p} f_j(x_i^{(j)}) + \varepsilon_i, \qquad i = 1, \ldots, n,$$

where $c$ is the intercept term, $\varepsilon_i$ are i.i.d. random variables with mean zero and $f_j : \mathbb{R} \to \mathbb{R}$ are smooth univariate functions. For identification purposes, we assume that all $f_j$ are centered, that is,

$$\sum_{i=1}^{n} f_j(x_i^{(j)}) = 0$$

for $j = 1, \ldots, p$. We consider the case of fixed design, that is, we treat the predictors $x^{(1)}, \ldots, x^{(p)}$ as nonrandom.

With some slight abuse of notation we also denote by $f_j$ the $n$-dimensional vector $(f_j(x_1^{(j)}), \ldots, f_j(x_n^{(j)}))^T$. For a vector $f \in \mathbb{R}^n$, we define $\|f\|_n^2 = \frac{1}{n} \sum_{i=1}^{n} f_i^2$.

2.1. *The sparsity-smoothness penalty.* In order to construct an estimator which encourages sparsity at the function level, penalizing the norms $\|f_j\|_n$ would be a suitable approach. Some theory for the case where a truncated orthogonal basis with $O(n^{1/5})$ basis functions for each component $f_j$ is used has been developed in [26].

If we use a large number of basis functions, which is necessary to be able to capture some functions at high complexity, the resulting estimator will produce function estimates which are too wiggly if the underlying true functions are very smooth. Hence, we need some additional control or restrictions of the smoothness of the estimated functions. In order to get sparse and sufficiently smooth function estimates, we propose the sparsity-smoothness penalty

$$J(f_j) = \lambda_1 \sqrt{\|f_j\|_n^2 + \lambda_2 I^2(f_j)},$$

where

$$I^2(f_j) = \int (f_j''(x))^2 \, dx$$

measures the smoothness of $f_j$. The two tuning parameters $\lambda_1, \lambda_2 \geq 0$ control the amount of penalization.

Our estimator is given by the following penalized least squares problem:

$$\hat{f}_1, \ldots, \hat{f}_p = \underset{f_1, \ldots, f_p \in \mathcal{F}}{\arg\min} \left\| Y - \sum_{j=1}^{p} f_j \right\|_n^2 + \sum_{j=1}^{p} J(f_j), \tag{1}$$

where $\mathcal{F}$ is a suitable class of functions and $Y = (Y_1, \ldots, Y_n)^T$ is the vector of responses. We assume the same level of regularity for each function $f_j$. If $Y$ is centered, we can omit an unpenalized intercept term and the nature of the objective function in (1) automatically forces the function estimates $\hat{f}_1, \ldots, \hat{f}_p$ to be centered.

PROPOSITION 1. *Let $a, b \in \mathbb{R}$ such that $a < \min_{i,j}\{x_i^{(j)}\}$ and $b > \max_{i,j}\{x_i^{(j)}\}$. Let $\mathcal{F}$ be the space of functions that are twice continuously differentiable on $[a, b]$ and assume that there exist minimizers $\hat{f}_j \in \mathcal{F}$ of (1). Then the $\hat{f}_j$'s are natural cubic splines with knots at $x_i^{(j)}, i = 1, \ldots, n$.*

A proof is given in Appendix A. Hence, we can restrict ourselves to the finite-dimensional space of natural cubic splines instead of considering the infinite-dimensional space of twice continuously differentiable functions.

In the following subsection, we illustrate the existence and the computation of the estimator.

2.2. *Computational algorithm.* For each function $f_j$, we use a cubic B-spline parameterization with a reasonable amount of knots or basis functions. A typical choice would be to use $K - 4 \asymp \sqrt{n}$ interior knots that are placed at the empirical quantiles of $x^{(j)}$. Hence, we parameterize

$$f_j(x) = \sum_{k=1}^{K} \beta_{j,k} b_{j,k}(x),$$

where $b_{j,k} \colon \mathbb{R} \to \mathbb{R}$ are the B-spline basis functions and $\beta_j = (\beta_{j,1}, \ldots, \beta_{j,K})^T \in \mathbb{R}^K$ is the parameter vector corresponding to $f_j$. Based on the basis functions, we can construct an $n \times pK$ design matrix $B = [B_1|B_2|\cdots|B_p]$, where $B_j$ is the $n \times K$ design matrix of the B-spline basis of the $j$th predictor, that is, $B_{j,il} = b_{j,l}(x_i^{(j)})$.

For twice continuously differentiable functions, the optimization problem (1) can now be reformulated as

$$\hat{\beta} = \operatorname*{arg\,min}_{\beta=(\beta_1,\ldots,\beta_p)} \|Y - B\beta\|_n^2 + \lambda_1 \sum_{j=1}^{p} \sqrt{\frac{1}{n}\beta_j^T B_j^T B_j \beta_j + \lambda_2 \beta_j^T \Omega_j \beta_j}, \quad (2)$$

where the $K \times K$ matrix $\Omega_j$ contains the inner products of the second derivatives of the B-spline basis functions, that is,

$$\Omega_{j,kl} = \int b_{j,k}''(x) b_{j,l}''(x)\,dx$$

for $k, l \in \{1, \ldots, K\}$.

Hence, (2) can be rewritten as a general group lasso problem [32]

$$\hat{\beta} = \operatorname*{arg\,min}_{\beta=(\beta_1,\ldots,\beta_p)} \|Y - B\beta\|_n^2 + \lambda_1 \sum_{j=1}^{p} \sqrt{\beta_j^T M_j \beta_j}, \quad (3)$$

where $M_j = \frac{1}{n} B_j^T B_j + \lambda_2 \Omega_j$. By decomposing (e.g., using the Cholesky decomposition) $M_j = R_j^T R_j$ for some quadratic $K \times K$ matrix $R_j$ and by defining $\tilde{\beta}_j = R_j \beta_j$, $\tilde{B}_j = B_j R_j^{-1}$, (3) reduces to

$$\hat{\tilde{\beta}} = \operatorname*{arg\,min}_{\tilde{\beta}=(\tilde{\beta}_1,\ldots,\tilde{\beta}_p)} \|Y - \tilde{B}\tilde{\beta}\|_n^2 + \lambda_1 \sum_{j=1}^{p} \|\tilde{\beta}_j\|, \quad (4)$$

where $\|\tilde{\beta}_j\| = \sqrt{K}\|\tilde{\beta}_j\|_K$ is the Euclidean norm in $\mathbb{R}^K$. This is an ordinary group lasso problem for any fixed $\lambda_2$, and hence the existence of a solution is guaranteed. For $\lambda_1$ large enough, some of the coefficient groups $\beta_j \in \mathbb{R}^K$ will be estimated to be exactly zero. Hence, the corresponding function estimate will be zero. Moreover, there exists a value $\lambda_{1,\max} < \infty$ such that $\hat{\tilde{\beta}}_1 = \cdots = \hat{\tilde{\beta}}_p = 0$ for $\lambda_1 \geq \lambda_{1,\max}$. This is especially useful to construct a grid of $\lambda_1$ candidate values for cross-validation (usually on the log-scale).

Regarding the uniqueness of the identified components, we have equivalent results as for the lasso. Define $S(\tilde{\beta}; \tilde{B}) = \|Y - \tilde{B}\tilde{\beta}\|_n^2$. Similar to [25], we have the following proposition.

PROPOSITION 2. *If $pK \leq n$, and if $\tilde{B}$ has full rank, a unique solution of (4) exists. If $pK > n$, there exists a convex set of solutions of (4). Moreover, if $\|\nabla_{\tilde{\beta}_j} S(\hat{\tilde{\beta}}; \tilde{B})\| < \lambda_1$, then $\hat{\tilde{\beta}}_j = 0$ and all other solutions $\hat{\tilde{\beta}}_{\text{other}}$ satisfy $\hat{\tilde{\beta}}_{\text{other},j} = 0$.*

A proof can be found in Appendix A.

By rewriting the original problem (1) in the form of (4), we can make use of already existing algorithms [16, 21, 32] to compute the estimator.

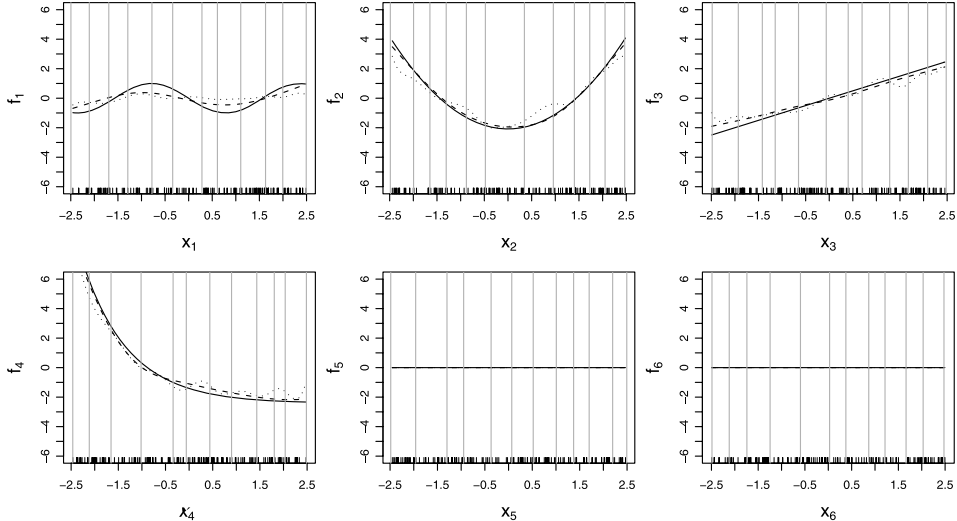

FIG. 1. *True functions $f_j$ (solid) and estimated functions $\hat{f}_j$ (dashed) for the first 6 components of a simulation run of Example 1 in Section 3. Small vertical bars indicate original data and grey vertical lines knot positions. The dotted lines are the function estimates when no smoothness penalty is used, that is, when setting $\lambda_2 = 0$.*

Coordinate-wise approaches as in [21, 32] are efficient and have rigorous convergence properties. Thus, we are able to compute the estimator exactly, even if $p$ is very large.

An example of estimated functions, from simulated data according to Example 1 in Section 3, is shown in Figure 1. For illustrational purposes, we have also plotted the estimator which involves no smoothness penalty ($\lambda_2 = 0$). The latter clearly shows that for this example, the function estimates are "too wiggly" compared to the true functions. As we will also see later, the smoothness penalty plays a key role for the theory.

REMARK 1. Alternative possibilities of our penalty would be to use either (i) $J(f_j) = \lambda_1 \|f_j\|_n + \lambda_2 I(f_j)$ or (ii) $J(f_j) = \lambda_1 \|f_j\|_n + \lambda_2 I^2(f_j)$. Both approaches lead to a sparse estimator. While proposal (i) also enjoys nice theoretical properties (see also Section 5.2), it is computationally more demanding, because it leads to a second order cone programming problem. Proposal (ii) basically leads again to a group lasso problem, but appears to have theoretical drawbacks, that is, the term $\lambda_2 I^2(f_j)$ is really needed within the square root.

2.3. *Oracle results.* We present now an oracle inequality for the penalized estimator. The proofs can be found in Appendix A.

For the theoretical analysis, we introduce an additional penalty parameter $\lambda_3 \geq 0$ for technical reasons. We consider, here, a penalty of the form

$$J(f_j) = \lambda_1 \sqrt{\|f_j\|_n^2 + \lambda_2 I^2(f_j)} + \lambda_3 I^2(f_j).$$

This penalty involves three smoothing parameters $\lambda_1$, $\lambda_2$ and $\lambda_3$. One may reduce this to a single smoothing parameter by choosing

$$\lambda_2 = \lambda_3 = \lambda_1^2,$$

(see Theorem 1 below). In the simulations however, the choice $\lambda_3 = 0$ turned out to provide slightly better results than the choice $\lambda_2 = \lambda_3$. With $\lambda_3 = 0$, the theory goes through provided the smoothness $I(\hat{f}_j)$ remains bounded in an appropriate sense.

We let $f^0$ denote the "true" regression function (which is not necessarily additive), that is, we suppose the regression model

$$Y_i = f^0(x_i) + \varepsilon_i,$$

where $x_i = (x_i^{(1)}, \ldots, x_i^{(p)})^T$ for $i = 1, \ldots, n$, and where $\varepsilon_1, \ldots, \varepsilon_n$ are independent random errors with $\mathbb{E}[\varepsilon_i] = 0$. Let $f^*$ be a (sparse) additive approximation of $f^0$ of the form

$$f^*(x_i) = c^* + \sum_{j=1}^{p} f_j^*(x_i^{(j)}),$$

where we take $c^* = \mathbb{E}[\bar{Y}]$, $\bar{Y} = \sum_{i=1}^{n} Y_i/n$. The result of this subsection (Theorem 1) holds for any such $f^*$ satisfying the compatibility condition below. Thus, one may invoke the optimal additive predictor among such $f^*$, which we will call the "oracle." For an additive function $f$, the squared distance $\|f - f^0\|_n^2$ can be decomposed into

$$\|f - f^0\|_n^2 = \|f - f_{\text{add}}^0\|_n^2 + \|f_{\text{add}}^0 - f^0\|_n^2,$$

where $f_{\text{add}}^0$ is the projection of $f^0$ on the space of additive functions. Thus, when $f^0$ is itself not additive, the oracle can be seen as the best sparse approximation of the projection $f_{\text{add}}^0$ of $f^0$.

The *active set* is defined as

(5) $$\mathcal{A}_* = \{j : \|f_j^*\|_n \neq 0\}.$$

We define, for $j = 1, \ldots, p$,

$$\tau_n^2(f_j) = \|f_j\|_n^2 + \lambda^{2-\gamma} I^2(f_j).$$

Moreover, we let $0 < \eta < 1$ be some fixed value. The constant $4/(1-\eta)$ appearing below in the compatibility condition is stated in this form to facilitate reference, later in the proof of Theorem 1.

We will use a compatibility condition, in the spirit of the incoherence conditions used for proving oracle inequalities for the standard lasso (see, e.g., [3, 8, 9, 10, 30]). To avoid digressions, we will not attempt to formulate the most general condition. A discussion can be found in Section 5.1.

COMPATIBILITY CONDITION. For some constants $0 < \eta < 1$ and $0 < \phi_{n,*} \leq 1$, and for all $\{f_j\}_{j=1}^p$ satisfying

$$\sum_{j=1}^p \tau_n(f_j) \leq \frac{4}{1-\eta} \sum_{j \in \mathcal{A}_*} \tau_n(f_j),$$

the following inequality is met:

$$\sum_{j \in \mathcal{A}_*} \|f_j\|_n^2 \leq \left( \left\| \sum_{j=1}^p f_j \right\|_n^2 + \lambda^{2-\gamma} \sum_{j \in \mathcal{A}_*} I^2(f_j) \right) \bigg/ \phi_{n,*}^2.$$

For practical applications, the compatibility condition cannot be checked because the set $\mathcal{A}_*$ is unknown.

Consider the general case where $I$ is some semi-norm, for example, as in Section 2.1. For mathematical convenience, we write

(6) $$f_j = g_j + h_j$$

with $g_j$ and $h_j$ centered and orthogonal functions, that is,

$$\sum_{i=1}^n g_{j,i} = \sum_{i=1}^n h_{j,i} = 0$$

and

$$\sum_{i=1}^n g_{j,i} h_{j,i} = 0,$$

such that $I(h_j) = 0$ and $I(g_j) = I(f_j)$. The functions $h_j$ are assumed to lie in a $d$-dimensional space. The entropy of $(\{g_j : I(g_j) = 1\}, \|\cdot\|_n)$ is denoted by $H_j(\cdot)$; see, for example, [29]. We assume that for all $j$,

(7) $$H_j(\delta) \leq A\delta^{-2(1-\alpha)}, \qquad \delta > 0,$$

where $0 < \alpha < 1$ and $A > 0$ are constants. When $I^2(f_j) = \int (f_j''(x))^2 \, dx$, the functions $h_j$ are the linear part of $f_j$, that is, $d = 1$. Moreover, one then has $\alpha = 3/4$ (see, e.g., [29], Lemma 3.9).

Finally, we assume sub-Gaussian tails for the errors: for some constants $L$ and $M$,

(8) $$\max_i \mathbb{E}[\exp(\varepsilon_i^2/L)] \leq M.$$

The next lemma presents the behavior of the empirical process. We use the notation $(\varepsilon, f)_n = \frac{1}{n}\sum_{i=1}^n \varepsilon_i f(x_i)$ for the inner product. Define

(9) $$\mathcal{S} = \mathcal{S}_1 \cap \mathcal{S}_2 \cap \mathcal{S}_3,$$

where

$$\mathcal{S}_1 = \left\{\max_j \sup_{g_j}\left(\frac{2|(\varepsilon, g_j)_n|}{\|g_j\|_n^\alpha I^{1-\alpha}(g_j)}\right) \leq \xi_n\right\},$$

$$\mathcal{S}_2 = \left\{\max_j \sup_{h_j}\left(\frac{2|(\varepsilon, h_j)_n|}{\|h_j\|_n}\right) \leq \xi_n\right\}$$

and

$$\mathcal{S}_3 = \{\bar{\varepsilon} \leq \xi_n\}, \qquad \bar{\varepsilon} = \frac{1}{n}\sum_{i=1}^n \varepsilon_i.$$

For an appropriate choice of $\xi_n$, the set $\mathcal{S}$ has large probability.

LEMMA 1. *Assume (7) and (8). There exist constants $c$ and $C$ depending only on $d$, $\alpha$, $A$, $L$ and $M$, such that for*

$$\xi_n \geq C\sqrt{\frac{\log p}{n}},$$

*one has*

$$\mathbb{P}(\mathcal{S}) \geq 1 - c\exp[-n\xi_n^2/c^2].$$

For $\alpha \in (0,1)$, we define its "conjugate" $\gamma = 2(1-\alpha)/(2-\alpha)$. Recall that when $I^2(f_j) = \int (f_j''(x))^2\,dx$, one has $\alpha = 3/4$, and hence $\gamma = 2/5$.

We are now ready to state the oracle result for $\hat{f} = \hat{c} + \sum_{j=1}^p \hat{f}_j$ as defined in (1), with $\hat{c} = \bar{Y}$.

THEOREM 1. *Suppose the compatibility condition is met. Take for $j = 1, \ldots, p$,*

$$J(f_j) = \lambda_1\sqrt{\|f_j\|_n^2 + \lambda_2 I^2(f_j)} + \lambda_3 I^2(f_j)$$

*with $\lambda_1 = \lambda^{(2-\gamma)/2}$ and $\lambda_2 = \lambda_3 = \lambda_1^2$, and with $\xi_n\sqrt{2}/\eta \leq \lambda \leq 1$. Then on the set $\mathcal{S}$ given in (9), it holds that*

$$\|\hat{f} - f_{\text{add}}^0\|_n^2 + 2(1-\eta)\lambda^{(2-\gamma)/2}\sum_{j=1}^p \tau_n(\hat{f}_j - f_j^*) + \lambda^{2-\gamma}\sum_{j=1}^p I^2(\hat{f}_j)$$

$$\leq 3\|f^* - f_{\text{add}}^0\|_n^2 + 3\lambda^{2-\gamma}\sum_{j \in \mathcal{A}_*}\left[I^2(f_j^*) + \frac{8}{\phi_{n,*}^2}\right] + 2\xi_n^2.$$

The result of Theorem 1 does not depend on the number of knots (basis functions) which are used to build the functions $\hat{f}_j$, as long as $\hat{f}_j$ and $\hat{f}_j^*$ use the same basis functions.

We would like to point out that the theory of Theorem 1 goes through with only two tuning parameters $\lambda_1$ and $\lambda_2$, but with the additional restriction that $I(\hat{f}_j)$ is appropriately bounded.

We also remark that we did not attempt to optimize the constants given in Theorem 1, but rather looked for a simple explicit bound.

REMARK 2. Assume that $\phi_{n,*}$ is bounded away from zero. For example, this holds with large probability for a realization of a design with independent components (see Section 5.1). In view of Lemma 1, one may take (under the conditions of this lemma) the smoothing parameter $\lambda$ of order $\sqrt{\log p/n}$. For $I^2(f_j) = \int (f_j''(x))^2\, dx$, $\gamma = 2/5$ and this gives $\lambda^{2-\gamma}$ of order $(\log p/n)^{4/5}$, which is up to the log-term the usual rate for estimating a twice differentiable function. If the oracle $f^*$ has bounded smoothness $I(f_j^*)$ for all $j$, Theorem 1 yields the convergence rate $p_{\text{act}}(\log p/n)^{4/5}$, with $p_{\text{act}} = |\mathcal{A}_*|$ being the number of active variables the oracle needs. This is again up to the log-term, the same rate one would obtain if it was known beforehand which of the $p$ functions are relevant. For general $\phi_{n,*}$, we have the convergence rate $p_{\text{act}}\phi_{n,*}^{-2}(\log p/n)^{4/5}$.

Furthermore, the result implies that with large probability, the estimator selects a sup-set of the active functions, provided that the latter have enough signal (such kind of variable screening results have been established for the lasso in linear and generalized linear models [24, 30]). More precisely, we have the following corollary.

COROLLARY 1. *Let $\mathcal{A}_0 = \{j : \|f_{\text{add},j}^0\|_n \neq 0\}$ be the active set of $f_{\text{add}}^0$. Assume the compatibility condition holds for $\mathcal{A}_0$, with constant $\phi_{n,0}$. Suppose also that for $j \in \mathcal{A}_0$, the smoothness is bounded, say $I(f_{\text{add},j}^0) \leq 1$. Choosing $f^* = f_{\text{add}}^0$ in Theorem 1, tells us that on $\mathcal{S}$,*

$$\sum_{j=1}^{p} \|\hat{f}_j - f_{\text{add},j}^0\|_n \leq C\lambda^{(2-\gamma)/2}|\mathcal{A}_0|/\phi_{n,0}^2 + 2\xi_n^2$$

*for some constant $C$. Hence, if*

$$\|f_{\text{add},j}^0\|_n > C\lambda^{(2-\gamma)/2}|\mathcal{A}_0|/\phi_{n,0}^2 + 2\xi_n^2, \qquad j \in \mathcal{A}_0,$$

*we have (on $\mathcal{S}$), that the estimated active set $\{j : \|\hat{f}_j\|_n \neq 0\}$ contains $\mathcal{A}_0$.*

2.4. *Comparison with related results.* After an earlier version of this paper, similar results have been published in [17]. Here, we point out some differences and similarities between our work and [17].

In [17], the framework of reproducing kernel hilbert spaces (RKHS) is considered, as for example, used in COSSO [19], while we use penalties based on smoothness seminorms. Hence, the two frameworks are rather different, at least from a mathematical point of view. The results in [17] are valid for a large class of loss functions, although we would like to point out that the quadratic loss as studied here is not covered in [17] since they assume that the loss function is appropriately bounded.

The oracle result and the conditions in [17] are similar to our Theorem 1. Regarding the convergence rate (see Remark 2), the rates obtained in [17] are similar in spirit to ours. In [17], the rate is slower than ours if the "smoothness" $\beta$ is equal to 2. Moreover, "smoothness" in [17] is very much intertwined with the unknown distribution of the covariables, whereas in our work "smoothness" is defined, for example, in terms of Sobolev-norms.

Compared to the work in [17], and, for example, COSSO [19], we gain flexibility through the introduction of the additional penalty parameter $\lambda_2$ for (separately) controlling the smoothness. In addition, we present an algorithm in Section 2.2 which is efficient with mathematically established convergence results.

## 3. Numerical examples.

3.1. *Simulations.* In this section, we investigate the empirical properties of the proposed estimator. We compare our approach with the boosting approach of [7], where smoothing splines with low degrees of freedom are used as base learners; see also [5]. For $p=1$, boosting with splines is known to be able to adapt to the smoothness of the underlying true function [7]. Generally, boosting is a very powerful machine learning method and a wide variety of software implementations are available, for example, the R add-on package mboost.

We use a training set of $n$ samples to train the different methods. An independent validation set of size $\lfloor n/2 \rfloor$ is used to select the prediction optimal tuning parameters $\lambda_1$ and $\lambda_2$. We use grids (on the log-scale) for both $\lambda_1$ and $\lambda_2$, where the grid for $\lambda_1$ is of size 100 and the grid for $\lambda_2$ is typically of about size 15. For boosting, the number of boosting iterations is used as tuning parameter. The shrinkage factor $\nu$ and the degrees of freedom $df$ of the boosting procedure are set to their default values $\nu = 0.1$ and $df = 4$; see also [5].

By SNR, we denote the signal-to-noise ratio, which is defined as

$$\text{SNR} = \frac{\text{Var}(f(X))}{\text{Var}(\varepsilon)},$$

where $f = f^0 : \mathbb{R}^p \to \mathbb{R}$ is the true underlying function.

A total of 100 simulation runs are used for each of the following settings.

3.1.1. *Models.* We use the following simulation models.

EXAMPLE 1 ($n = 150$, $p = 200$, $p_{\text{act}} = 4$, SNR $\approx 15$). This example is similar to Example 1 in [26] and [15]. The model is

$$Y_i = f_1(x_i^{(1)}) + f_2(x_i^{(2)}) + f_3(x_i^{(3)}) + f_4(x_i^{(4)}) + \varepsilon_i, \qquad \varepsilon_i \text{ i.i.d. } N(0,1),$$

with

$$f_1(x) = -\sin(2x), \qquad f_2(x) = x_2^2 - 25/12, \qquad f_3(x) = x,$$
$$f_4(x) = e^{-x} - 2/5 \cdot \sinh(5/2).$$

The covariates are simulated from independent Uniform$(-2.5, 2.5)$ distributions. The true and the estimated functions of a simulation run are illustrated in Figure 1.

EXAMPLE 2 ($n = 100$, $p = 1000$, $p_{\text{act}} = 4$, SNR $\approx 6.7$). As above but high dimensional and correlated. The covariates are simulated according to a multivariate normal distribution with covariance matrix $\Sigma_{ij} = 0.5^{|i-j|}; i, j = 1, \ldots, p$.

EXAMPLE 3 [$n = 100$, $p = 80$, $p_{\text{act}} = 4$, SNR $\approx 9$ ($t = 0$), $\approx 7.9$ ($t = 1$)]. This is similar to Example 1 in [19] but with more predictors. The model is

$$Y_i = 5f_1(x_i^{(1)}) + 3f_2(x_i^{(2)}) + 4f_3(x_i^{(3)}) + 6f_4(x_i^{(4)}) + \varepsilon_i, \qquad \varepsilon_i \text{ i.i.d. } N(0, 1.74),$$

with

$$f_1(x) = x, \qquad f_2(x) = (2x-1)^2, \qquad f_3(x) = \frac{\sin(2\pi x)}{2 - \sin(2\pi x)}$$

and

$$f_4(x) = 0.1 \sin(2\pi x) + 0.2 \cos(2\pi x) + 0.3 \sin^2(2\pi x)$$
$$+ 0.4 \cos^3(2\pi x) + 0.5 \sin^3(2\pi x).$$

The covariates $x = (x^{(1)}, \ldots, x^{(p)})^T$ are simulated according to

$$x^{(j)} = \frac{W^{(j)} + tU}{1+t}, \qquad j = 1, \ldots, p,$$

where $W^{(1)}, \ldots, W^{(p)}$ and $U$ are i.i.d. Uniform$(0, 1)$. For $t = 0$ this is the independent uniform case. The case $t = 1$ results in a design with correlation 0.5 between all covariates.

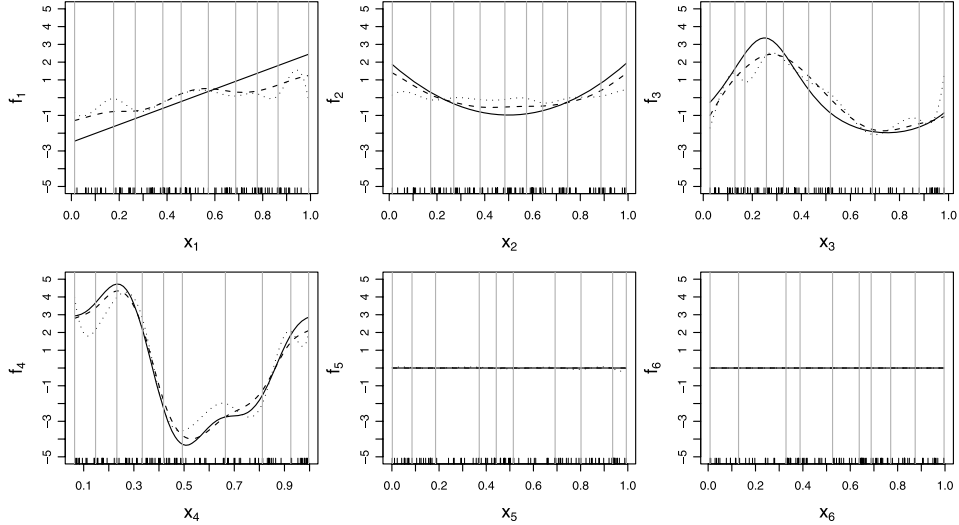

FIG. 2. *True functions $f_j$ (solid) and estimated functions $\hat{f}_j$ (dashed) for the first 6 components of a simulation run of Example 3 ($t = 0$). Small vertical bars indicate original data and grey vertical lines knot positions. The dotted lines are the function estimates when no smoothness penalty is used, that is, when setting $\lambda_2 = 0$.*

The true functions and the first 6 estimated functions of a simulation run with $t = 0$ are illustrated in Figure 2.

Moreover, we also consider a "high-frequency" situation where we use $f_3(8x)$ and $f_4(4x)$ instead of $f_3(x)$ and $f_4(x)$. The corresponding signal-to-noise ratios for these models are SNR $\approx 9$ for $t = 0$ and SNR $\approx 8.1$ for $t = 1$.

EXAMPLE 4 [$n = 100$, $p = 60$, $p_{\text{act}} = 12$, SNR $\approx 9$ ($t = 0$), $\approx 11.25$ ($t = 1$)]. This is similar to Example 2 in [19] but with fewer observations. We use the same functions as in Example 3. The model is

$$Y_i = f_1(x_i^{(1)}) + f_2(x_i^{(2)}) + f_3(x_i^{(3)}) + f_4(x_i^{(4)})$$
$$+ 1.5 f_1(x_i^{(5)}) + 1.5 f_2(x_i^{(6)}) + 1.5 f_3(x_i^{(7)}) + 1.5 f_4(x_i^{(8)})$$
$$+ 2 f_1(x_i^{(9)}) + 2 f_2(x_i^{(10)}) + 2 f_3(x_i^{(11)}) + 2 f_4(x_i^{(12)}) + \varepsilon_i$$

with $\varepsilon_i$ i.i.d. $N(0, 0.5184)$. The covariates are simulated as in Example 3.

3.1.2. *Performance measures.* In order to compare the prediction performances, we use the mean squared prediction error

$$\text{PE} = \mathbb{E}_X[(\hat{f}(X) - f(X))^2]$$

TABLE 1
*Results of the different simulation models. Reported is the mean of the ratio of the prediction error of the two methods. SSP: sparsity-smoothness penalty approach, boost: boosting with smoothing splines. Standard deviations are given in parentheses*

| Model | $PE_{SSP}/PE_{boost}$ |
|---|---|
| Example 1 | 0.93 (0.13) |
| Example 2 | 0.96 (0.10) |
| Example 3 ($t=0$) | 0.81 (0.13) |
| Example 3 ($t=1$) | 0.90 (0.19) |
| Example 3 "high-freq" ($t=0$) | 0.65 (0.11) |
| Example 3 "high-freq" ($t=1$) | 0.57 (0.10) |
| Example 4 ($t=0$) | 0.89 (0.10) |
| Example 4 ($t=1$) | 0.88 (0.13) |

as performance measure. The above expectation is approximated by a sample of 10,000 points from the distribution of $X$. In each simulation run, we compute the ratio of the prediction performance of the two methods. Finally, we take the mean of the ratios over all simulation runs.

For variable selection properties, we use the number of true positives (TP) and false positives (FP) at each simulation run. We report the average number over all runs to compare the different methods.

3.1.3. *Results.* The results are summarized in Tables 1 and 2. The sparsity-smoothness penalty approach (SSP) has smaller prediction error than boosting, especially for the "high-frequency" situations. Because the weak learners of the boosting method only use 4 degrees of freedom, boosting tends to neglect or underestimate those components with higher oscillation. This can also be observed with respect to the number of true positives. By relaxing the smoothness penalty (i.e., choosing $\lambda_2$ small or setting $\lambda_2 = 0$), SSP is able to handle the high-frequency situations, at the cost of too wiggly function estimates for the remaining components. Using a different amount of regularization for sparsity and smoothness, SSP can work with a large amount of basis functions in order to be flexible enough to capture sophisticated functional relationships and, on the other side, to produce smooth estimates if the underlying functions are smooth.

With the exception of the high-frequency examples, the number of true positives (TP) is very similar for both methods. There is no clear trend with respect to the number of false positives (FP).

3.2. *Real data.* In this section, we would like to compare the different estimators on real data sets.

3.2.1. *Tecator.* The `meatspec` data set contains data from the Tecator Infratec Food and Feed Analyzer. It is, for example, available in the R add-on package `faraway` and on StatLib. The $p = 100$ predictors are channel spectrum measurements, and are therefore highly correlated. A total of $n = 215$ observations are available.

The data is split into a training set of size 100 and a validation set of size 50. The remaining data are used as test set. On the training dataset, the first 30 principal components are calculated, scaled to unit variance and used as covariates in additive modeling. Moreover, the validation and test data sets are transformed to correspond to the principal components of the training data set. We fit an additive model to predict the logarithm of the fat content. This is repeated 50 times. For each split into training and test data, we compute the ratio of the prediction errors from the SSP and boosting method on the test data, as in Section 3.1.2. The mean of the ratio over the 50 splits is 0.86, the corresponding standard deviation is 0.46. This indicates superiority of our sparsity-smoothness penalty approach.

3.2.2. *Motif regression.* In motif regression problems [11], the aim is to predict gene expression levels or binding intensities based on information on the DNA sequence. For our specific dataset, from the Ricci lab at ETH Zurich, we have binding intensities $Y_i$ of a certain transcription factor (TF) at 287 regions on the DNA. Moreover, for each region $i$, motif scores $x_i^{(1)}, \ldots, x_i^{(p)}, p = 196$ are available. A motif is a candidate for the binding site of the TF on the DNA, typically a 5–15bp long DNA sequence. The score $x_i^{(j)}$ measures how well the $j$th motif is represented in the $i$th region. The candidate list of motifs and their corresponding scores were created with a variant of the MDScan algorithm [20]. The main goal here is to find the relevant covariates.

TABLE 2
*Average values of the number of true (TP) and false (FP) positives. Standard deviations are given in parentheses*

| Model | $TP_{SSP}$ | $FP_{SSP}$ | $TP_{boost}$ | $FP_{boost}$ |
|---|---|---|---|---|
| Example 1 | 4.00 (0.00) | 24.30 (14.11) | 4.00 (0.00) | 22.18 (12.75) |
| Example 2 | 3.47 (0.61) | 34.37 (17.38) | 3.60 (0.64) | 28.76 (20.15) |
| Example 3 ($t=0$) | 4.00 (0.00) | 20.20 (9.30) | 4.00 (0.00) | 21.61 (10.90) |
| Example 3 ($t=1$) | 3.93 (0.29) | 19.28 (9.61) | 3.92 (0.27) | 18.65 (8.35) |
| Example 3 "high-freq" ($t=0$) | 2.80 (0.78) | 12.26 (7.61) | 2.16 (0.94) | 9.23 (9.74) |
| Example 3 "high-freq" ($t=1$) | 2.46 (0.85) | 11.17 (8.50) | 1.59 (1.27) | 13.24 (13.89) |
| Example 4 ($t=0$) | 11.69 (0.56) | 21.23 (6.85) | 11.68 (0.57) | 25.91 (9.43) |
| Example 4 ($t=1$) | 10.64 (1.15) | 19.78 (7.51) | 10.67 (1.25) | 23.76 (9.89) |

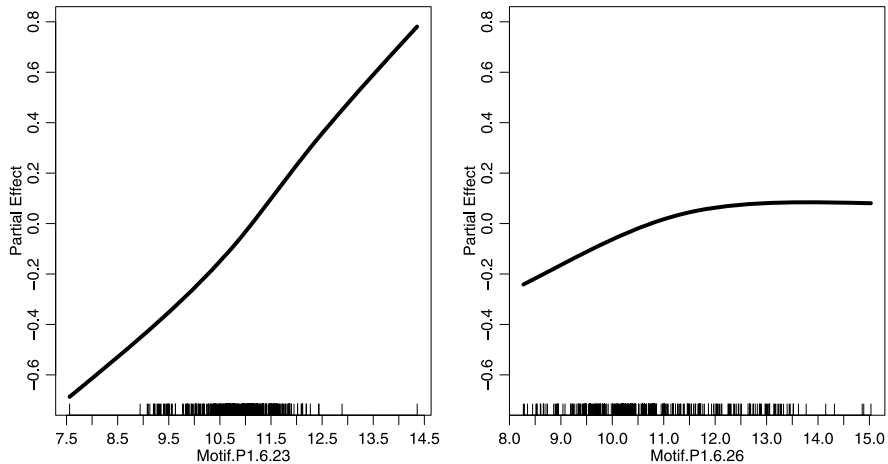

Fig. 3. *Estimated functions $\hat{f}_j$ of the two most stable motifs. Small vertical bar indicate original data.*

We used 5 fold cross-validation to determine the prediction optimal tuning parameters, yielding 28 active functions. To assess the stability of the estimated model, we performed a nonparametric bootstrap analysis. At each of the 100 bootstrap samples, we fit the model with the fixed optimal tuning parameters from above. The two functions which appear most often in the bootstrapped model estimates are depicted in Figure 3. While the left-hand side plot shows an approximate linear relationship, the effect of the other motif seems to diminish for larger values. Indeed, `Motif.P1.6.26` is the true (known) binding site. A follow-up experiment showed that the TF does not directly bind to `Motif.P1.6.23`. Hence, this motif is a candidate for a binding site of a co-factor (another TF) and needs further experimental validation.

## 4. Extensions.

4.1. *Generalized additive models.* Conceptually, we can also apply the sparsity-smoothness penalty from Section 2 to generalized linear models (GLM) by replacing the residual sum of squares $\|Y - \sum_{j=1}^p f_j\|_n^2$ by the corresponding negative log-likelihood function. We illustrate the method for logistic regression where $Y \in \{0, 1\}$. The negative log-likelihood as a function of the linear predictor $\eta$ and the response vector $Y$ is

$$\ell(\eta, Y) = -\frac{1}{n} \sum_{i=1}^n [Y_i \eta_i - \log\{1 + \exp(\eta_i)\}],$$

TABLE 3
*Results of different model sizes $p$. Reported is the mean of the ratio of the prediction error of the two methods. SSP: sparsity-smoothness penalty approach, boost: boosting with smoothing splines. Standard deviations are given in parentheses*

| $p$ | $\mathrm{PE}_{\mathrm{SSP}}/\mathrm{PE}_{\mathrm{boost}}$ |
|---|---|
| 250 | 0.93 (0.06) |
| 500 | 0.96 (0.07) |
| 1000 | 0.98 (0.05) |

where $\eta_i = c + \sum_{j=1}^{p} f_j(x_i^{(j)})$. The estimator is defined as

$$(10) \quad \hat{c}, \hat{f}_1, \ldots, \hat{f}_p = \underset{c \in \mathbb{R}, f_1, \ldots, f_p \in \mathcal{F}}{\arg\min} \ell\left(c + \sum_{j=1}^{p} f_j, Y\right) + \sum_{j=1}^{p} J(f_j).$$

This has a similar form as (1) with the exception that we have to explicitly include a (nonpenalized) intercept term $c$. Using the same arguments as in Section 2, leads to the fact that for twice continuously differentiable functions, the solution can be represented as a natural cubic spline and that (10) leads again to a group lasso problem. This can, for example, be minimized with the algorithm of [21]. We illustrate the performance of the estimator in a small simulation study.

4.1.1. *Small simulation study.* Denote by $f : \mathbb{R}^p \to \mathbb{R}$ the true function of Example 2 in Section 3. We simulate the the linear predictor $\eta$ as

$$\eta(X) = 1.5 \cdot (2 + f(X)),$$

where $X \in \mathbb{R}^p$ has the same distribution as in Example 2. The binary response $Y$ is then generated according to a Bernoulli distribution with probability $1/(1 + \exp(-\eta(X)))$, which results in a Bayes risk of approximately 0.17. The sample size $n$ is set to 100. The results for various model sizes $p$ are reported in Tables 3 and 4. The performance of the two methods is quite similar. SSP has a slightly lower prediction error. Regarding model selection properties, SSP has fewer false positives at the cost of slightly fewer true positives.

4.2. *Adaptivity.* Similar to the adaptive lasso [34], we can also use different penalties for the different components, that is, use a penalty of the form

$$J(f_j) = \lambda_1 \sqrt{w_{1,j} \|f_j\|_n^2 + \lambda_2 w_{2,j} I^2(f_j)},$$

where the weights $w_{1,j}$ and $w_{2,j}$ are ideally chosen in a data-adaptive way. If an initial estimator $\hat{f}_{j,\text{init}}$ is available, a choice would be to use

$$w_{1,j} = \frac{1}{\|\hat{f}_{j,\text{init}}\|_n^\gamma}, \qquad w_{2,j} = \frac{1}{I(\hat{f}_{j,\text{init}})^\gamma}$$

for some $\gamma > 0$. The estimator can then be computed similarly as described in Section 2.2. This allows for different degrees of smoothness for different components.

We have applied the adaptive estimator to the simulation models of Section 3. In each simulation run, we use weights (with $\gamma = 1$) based on the ordinary sparsity-smoothness estimator. For comparison, we compute the ratio of the prediction error of the adaptive and the ordinary sparsity-smoothness estimator at each simulation run. The results are summarized in Table 5. Both the prediction error and the number of false positives can be decreased by a good margin in all examples. The number of true positives gets slightly decreased in some examples.

## 5. Mathematical theory.

TABLE 4
*Average values of the number of true (TP) and false (FP) positives. Standard deviations are given in parentheses*

| $p$ | $\text{TP}_{\text{SSP}}$ | $\text{FP}_{\text{SSP}}$ | $\text{TP}_{\text{boost}}$ | $\text{FP}_{\text{boost}}$ |
|---|---|---|---|---|
| 250  | 2.94 (0.71) | 22.81 (10.56) | 3.09 (0.78) | 29.67 (14.91) |
| 500  | 2.56 (0.82) | 24.92 (12.47) | 2.80 (0.82) | 31.41 (17.28) |
| 1000 | 2.36 (0.84) | 26.45 (14.88) | 2.61 (0.71) | 33.69 (19.54) |

TABLE 5
*Results of the different simulation models. Reported is the mean of the ratio of the prediction error of the two methods and the average values of the number of true (TP) and false (FP) positives. SSP; adapt: adaptive sparsity-smoothness penalty approach, SSP: ordinary sparsity-smoothness penalty approach. Standard deviations are given in parentheses*

| Model | $\text{PE}_{\text{SSP;adapt}}/\text{PE}_{\text{SSP}}$ | TP | FP |
|---|---|---|---|
| Example 1 | 0.47 (0.13) | 4.00 (0.00) | 4.09 (4.63) |
| Example 2 | 0.63 (0.17) | 3.31 (0.71) | 6.12 (5.14) |
| Example 3 ($t=0$) | 0.53 (0.14) | 4.00 (0.00) | 4.64 (4.52) |
| Example 3 ($t=1$) | 0.63 (0.22) | 3.81 (0.46) | 5.04 (4.82) |
| Example 3 "high-freq" ($t=0$) | 0.87 (0.09) | 2.28 (0.78) | 2.98 (2.76) |
| Example 3 "high-freq" ($t=1$) | 0.91 (0.10) | 1.69 (0.73) | 2.59 (3.30) |
| Example 4 ($t=0$) | 0.77 (0.11) | 11.21 (0.84) | 8.18 (5.04) |
| Example 4 ($t=1$) | 0.88 (0.12) | 9.73 (1.29) | 7.93 (5.35) |

5.1. *On the compatibility condition.* We show in this subsection that the compatibility condition holds under reasonable conditions when

$$I(f_j) = \sqrt{\int_0^1 |f^{(s)}(x)|^2 \, dx}$$

is the Sobolev norm ($f_j^{(s)}$ being the $s$th derivative of $f_j$), and when in addition, the $X_i = (X_i^{(1)}, \ldots, X_i^{(p)})$ are i.i.d. copies of a $p$-dimensional random variable $X \in [0,1]^p$ with distribution $Q$. Then, the compatibility condition may be replaced by a theoretical variant, where the norm $\|\cdot\|_n$ is replaced by the theoretical $L_2(Q)$-norm $\|\cdot\|$. The theoretical compatibility condition (given below) is not about $n$-dimensional vectors, but about functions. In that sense, the sample size $n$ plays a less prominent role. For example, the theoretical compatibility condition is satisfied when the components $X^{(1)}, \ldots, X^{(p)}$ are independent.

The main assumption to make the replacement by a theoretical version possible, is the requirement that

$$\lambda^{1-\gamma}|\mathcal{A}_*|$$

[with $\gamma = 2/(2s+1)$] is small in an appropriate sense [see (11)]. This is comparable to the condition $\lambda|\mathcal{A}_*|$ being small, for the ordinary lasso (see, e.g., [9]). In fact, our approach for the transition from fixed to random design may also shed new light on the same transition for the lasso.

Let $X = (X^{(1)}, \ldots, X^{(p)}) \in [0,1]^p$ have distribution $Q$, and let $X_1, \ldots, X_n$ be i.i.d. copies of $X$. The marginal distribution of $X^{(j)}$ is denoted by $Q_j$. We write

$$\|f\|^2 = \int f^2 \, dQ$$

and for a function $f_j$ depending only on the $j$th variable $X^{(j)}$,

$$\|f_j\|^2 = \int f_j^2 \, dQ_j.$$

In this subsection, we assume all $f_j$'s are centered:

$$\int f_j \, dQ_j = 0, \qquad j = 1, \ldots, p.$$

Recall the notation

$$\tau_n^2(f_j) = \|f_j\|_n^2 + \lambda^{2-\gamma} I^2(f_j).$$

We now also define the theoretical counterparts

$$\tau^2(f_j) = \|f_j\|^2 + \lambda^{2-\gamma} I^2(f_j)$$

and write

$$\tau_{\text{tot}}(f) = \tau_{\text{in}}(f) + \tau_{\text{out}}(f),$$
$$\tau_{\text{in}}(f) = \sum_{j \in \mathcal{A}_*} \tau(f_j), \qquad \tau_{\text{out}}(f) = \sum_{j \notin \mathcal{A}_*} \tau(f_j).$$

One now may actually redress the proofs for the oracle inequality directly, in order to handle random design. This will generally lead to better constants as the approach that we now take, which is showing that the conditions for fixed design hold with large probability. The advantage of this detour is however that we do not have to repeat the main body of the proof.

The theoretical compatibility condition is of the same form as the empirical one, but with different constants.

THEORETICAL COMPATIBILITY CONDITION. For a constant $0 < \eta < 1$ and $0 < \phi_* \leq 1$, and for all $f$ satisfying

$$\tau_{\text{tot}}(f) \leq c_\eta \tau_{\text{in}}(f),$$

where

$$c_\eta = \frac{4(1+\eta)}{(1-\eta)^2},$$

we have

$$\sum_{j \in \mathcal{A}_*} \|f_j\|^2 \leq \left( \|f\|^2 + \lambda^{2-\gamma} \sum_{j \in \mathcal{A}_*} I^2(f_j) \right) \Big/ \phi_*^2.$$

Note that the theoretical compatibility condition trivially holds when the components of $X$ are independent. However, independence is not a necessary condition: much broader schemes are allowed.

Let $C_0$ be a constant and

$$\mathcal{S}_4 = \left\{ \sup_f \frac{|\|f\|_n^2 - \|f\|^2|}{\tau_{\text{tot}}^2(f)} \leq C_0 \lambda^{1-\gamma} \right\}.$$

In Appendix B, we show that for an appropriate value of $\lambda$, $\mathcal{S}_4$ has large probability, for a constant $C_0$ depending only on $s$, and on an assumed lower bound for the marginal densities of the $X^{(j)}$. In fact, it turns out that one can take $\lambda$ of order $\sqrt{\log p/n}$ under weak conditions, assuming $I(\cdot)$ is the Sobolev norm.

THEOREM 2. *Assume*

(11) $$\frac{2C_0 c_\eta^2 |\mathcal{A}_*| \lambda^{1-\gamma}}{\phi_*^2} \leq 1.$$

Then on $\mathcal{S}_4$, the theoretical compatibility condition implies the empirical one as given in Section 2.3, with constant

$$\frac{1}{\phi_{n,*}^2} = \left((1+\eta)(1+\phi_*^2) + \frac{2(1+\eta)}{\phi_*^2} + \eta\right).$$

As previously mentioned, condition (11) implies that the number of active components cannot grow too fast in order for $|\mathcal{A}_*|\lambda^{1-\gamma}$ being small.

We now have a quick closer look at the theoretical compatibility condition. The following two conditions are sufficient and might yield some more insight.

WELL-CONDITIONED ACTIVE SET CONDITION. We say that the active set $\mathcal{A}_*$ is well conditioned if for some constant $0 < \psi_* \le 1$, and for all $\{f_j\}_{j \in \mathcal{A}_*}$,

$$\sum_{j \in \mathcal{A}_*} \|f_j\|^2 \le \left\|\sum_{j \in \mathcal{A}_*} f_j\right\|^2 / \psi_*^2.$$

The inner product in $L_2(Q)$ between functions $f$ and $\tilde{f}$ is denoted by $(f, \tilde{f})$. No perfect canonical dependence in our setup amounts to the following condition.

NO PERFECT CANONICAL DEPENDENCE CONDITION. We say that the active and nonactive variables have no perfect canonical dependence, if for a constant $0 \le \rho_* < 1$, and all $\{f_j\}_{j=1}^p$, we have for $f_{\text{in}} = \sum_{j \in \mathcal{A}_*} f_j$ and $f_{\text{out}} = \sum_{j \notin \mathcal{A}_*} f_j$, that

$$\frac{|(f_{\text{in}}, f_{\text{out}})|}{\|f_{\text{in}}\|\|f_{\text{out}}\|} \le \rho_*.$$

The next lemma makes the link between the theoretical compatibility condition and the above two conditions.

LEMMA 2. *Let $f = f_{\text{in}} + f_{\text{out}}$ satisfy*

$$\frac{|(f_{\text{in}}, f_{\text{out}})|}{\|f_{\text{in}}\|\|f_{\text{out}}\|} \le \rho_* < 1.$$

*Then*

$$\|f_{\text{in}}\|^2 \le \|f\|^2 / (1 - \rho_*^2).$$

PROOF. Clearly,
$$\|f_{\rm in}\|^2 \leq \|f\|^2 + 2|(f_{\rm in}, f_{\rm out})| - \|f_{\rm out}\|^2.$$

Hence,
$$\|f_{\rm in}\|^2 \leq \|f\|^2 + 2\rho_*\|f_{\rm in}\|\|f_{\rm out}\| - \|f_{\rm out}\|^2 \leq \|f\|^2 + \rho_*^2\|f_{\rm in}\|^2. \qquad \square$$

COROLLARY 2. *A well-conditioned active set in combination with no perfect canonical dependence implies the theoretical compatibility condition with* $\phi_*^2 = \psi_*^2(1 - \rho_*^2)$.

REMARK 3. Canonical dependence is about the dependence structure of variables. To compare, let $X_{\rm in}$ and $X_{\rm out}$ be two random variables, with joint density $q$, and with marginal densities $q_{\rm in}$ and $q_{\rm out}$. Define for real-valued measurable functions $f_{\rm in}$ and $f_{\rm out}$, of $X_{\rm in}$ and $X_{\rm out}$, respectively, the squared norms $\|f_{\rm in}\|^2 = \int f_{\rm in}^2 q_{\rm in}$, and $\|f_{\rm out}\|^2 = \int f_{\rm out}^2 q_{\rm out}$, and the inner product $(f_{\rm in}, f_{\rm out}) = \int f_{\rm in} f_{\rm out} q$. Assume the functions are centered: $\int f_{\rm in} q_{\rm in} = \int f_{\rm out} q_{\rm out} = 0$. Suppose that for some constant $\rho_*$,
$$\int \frac{q^2}{q_{\rm in} q_{\rm out}} \leq 1 + \rho_*^2.$$

Then one can easily verify that $|(f_{\rm in}, f_{\rm out})| \leq \rho_*\|f_{\rm in}\|\|f_{\rm out}\|$. In other words, the *no perfect canonical dependence condition* is in this context the assumption that the density and the product density are, in $\chi^2$-sense, not too far off.

5.2. *On the choice of the penalty.* In this paper, we have chosen the penalty in such a way that it leads to good theoretical behavior (namely the oracle inequality of Theorem 1), as well as to computationally fast, and in fact already existing, algorithms. The penalty can be improved theoretically, at the cost of computational efficiency and simplicity.

Indeed, a main ingredient from the theoretical point of view is that the randomness of the problem (the behavior of the empirical process) should be taken care of. Let us recall Lemma 1 which says that the set $\mathcal{S}$ has large probability, and on $\mathcal{S}$ all functions $g_j$ satisfy
$$(\varepsilon, g_j)_n \leq \xi_n \|g_j\|_n^\alpha I^{1-\alpha}(g_j).$$

Our penalty was based on the inequality (which holds for any $a$ and $b$ positive)
$$a^\alpha b^{1-\alpha} \leq \sqrt{a^2 + b^2}.$$

More generally, it holds for any $q \geq 1$ that
$$a^\alpha b^{1-\alpha} \leq (a^q + b^q)^{1/q}.$$

In particular, the choice $q = 1$ would be a natural one, and would lead to an oracle inequality involving $I(f_j^*)$ instead of the square $I^2(f_j^*)$ on the right-hand side in Theorem 1. The penalty $\lambda^{(2-\gamma)/2} \sum_{j=1}^p \|f_j\|_n + \lambda^{2-\gamma} \sum_{j=1}^p I(f_j)$, corresponding to $q = 1$, still involves convex optimization but which is much more involved and hence less efficient to be solved; see also Remark 1 in Section 2.2.

One may also use the inequality

$$a^\alpha b^{1-\alpha} \leq a^2 + b^\gamma.$$

This leads to a "theoretically ideal" penalty of the from $\lambda^{2-\gamma} \sum_{j=1}^p I^\gamma(f_j) + \lambda \sum_{j=1}^p \|h_j\|_n$, where $h_j$ is from (6). It allows to adapt to small values of $I(f_j^*)$. But clearly, as this penalty is nonconvex, it may be computationally cumbersome. On the other hand, iterative approximations might prove to work well.

**6. Conclusions.** We present an estimator and algorithm for fitting sparse, high-dimensional generalized additive models. The estimator is based on a penalized likelihood. The penalty is new, as it allows for different regularization of the sparsity and the smoothness of the additive functions. It is exactly this combination which allows to derive oracle results for high-dimensional additive models. We also argue empirically that the inclusion of a smoothness-part into the penalty function yields much better results than having the sparsity-term only. Furthermore, we show that the optimization of the penalized likelihood can be written as a group lasso problem and hence, efficient coordinate-wise algorithms can be used which have provable numerical convergence properties.

We illustrate some empirical results for simulated and real data. Our new approach with the sparsity and smoothness penalty is never worse and sometimes substantially better than $L_2$-boosting for generalized additive model fitting [5, 7]. Furthermore, with an adaptive sparsity-smoothness penalty method, large additional performance gains are achieved. With the real data about motif regression for finding DNA-sequence motifs, one among two selected "stable" variables is known to be true, that is, it corresponds to a known binding site of a transcription factor.

## APPENDIX A: PROOFS

PROOF OF PROPOSITION 1. Because of the additive structure of $f$ and the penalty, it suffices to analyze each component $f_j, j = 1, \ldots, p$ independently. Let $\hat{f}_1, \ldots, \hat{f}_p$ be a solution of (1) and assume that some or all $\hat{f}_j$ are not natural cubic splines with knots at $x_i^{(j)}, i = 1, \ldots, n$. By Theorem 2.2 in

[13], we can construct natural cubic splines $\hat{g}_j$ with knots at $x_i^{(j)}, i=1,\ldots,n$ such that

$$\hat{g}_j(x_i^{(j)}) = \hat{f}_j(x_i^{(j)})$$

for $i=1,\ldots,n$ and $j=1,\ldots,p$. Hence,

$$\left\|Y - \sum_{j=1}^p \hat{g}_j\right\|_n^2 = \left\|Y - \sum_{j=1}^p \hat{f}_j\right\|_n^2$$

and

$$\|\hat{g}_j\|_n^2 = \|\hat{f}_j\|_n^2.$$

But by Theorem 2.3, in [13], $I^2(\hat{g}_j) \leq I^2(\hat{f}_j)$. Therefore, the value in the objective function (1) can be decreased. Hence, the minimizer of (1) must lie in the space of natural cubic splines. $\square$

PROOF OF PROPOSITION 2. The first part follows because of the strict convexity of the objective function. Consider now the case $pK > n$. The (necessary and sufficient) conditions for $\hat{\tilde{\beta}}$ to be a solution of the group lasso problem (4) are [32]

$$\|\nabla_{\tilde{\beta}_j} S(\hat{\tilde{\beta}}; \tilde{B})\| = \lambda_1 \qquad \text{for } \hat{\tilde{\beta}}_j \neq 0,$$

$$\|\nabla_{\tilde{\beta}_j} S(\hat{\tilde{\beta}}; \tilde{B})\| \leq \lambda_1 \qquad \text{for } \hat{\tilde{\beta}}_j = 0.$$

Assume that there exist two solutions $\hat{\tilde{\beta}}^{(1)}$ and $\hat{\tilde{\beta}}^{(2)}$ such that, for a component $j$, we have $\hat{\tilde{\beta}}_j^{(1)} = 0$ with $\|\nabla_{\tilde{\beta}_j} S(\hat{\tilde{\beta}}^{(1)}; \tilde{B})\| < \lambda_1$, but $\hat{\tilde{\beta}}_j^{(2)} \neq 0$. Because the set of all solutions is convex,

$$\hat{\tilde{\beta}}_\rho = (1-\rho)\hat{\tilde{\beta}}^{(1)} + \rho\hat{\tilde{\beta}}^{(2)}$$

is also a minimizer for all $\rho \in [0,1]$. By assumption $\hat{\tilde{\beta}}_{\rho,j} \neq 0$, and hence $\|\nabla_{\tilde{\beta}_j} S(\hat{\tilde{\beta}}_\rho; \tilde{B})\| = \lambda_1$ for all $\rho \in (0,1)$. Hence, it holds for $g(\rho) = \|\nabla_{\tilde{\beta}_j} S(\hat{\tilde{\beta}}_\rho; \tilde{B})\|$ that $g(0) < \lambda_1$ and $g(\rho) = \lambda_1$ for all $\rho \in (0,1)$. But this is a contradiction to the fact that $g(\cdot)$ is continuous. Hence, a nonactive (i.e., zero) component $j$ with $\|\nabla_{\tilde{\beta}_j} S(\hat{\tilde{\beta}}; \tilde{B})\| < \lambda_1$ cannot be active (i.e., nonzero) in any other solution. $\square$

**Proof of Lemma 1.** The result easily follows from Lemma 8.4 in [29], which we cite here for completeness.

LEMMA 3. *Let $\mathcal{G}$ be a collection of functions $g:\{x_1,\ldots,x_n\} \to \mathbb{R}$, endowed with a metric induced by the norm $\|g\|_n = (\frac{1}{n}\sum_{i=1}^{n} g^2(x_i))^{1/2}$. Let $H(\cdot)$ be the entropy of $\mathcal{G}$. Suppose that*

$$H(\delta) \leq A\delta^{-2(1-\alpha)} \qquad \forall \delta > 0.$$

*Furthermore, let $\varepsilon_1,\ldots,\varepsilon_n$ be independent centered random variables, satisfying*

$$\max_i \mathbb{E}[\exp(\varepsilon_i^2/L)] \leq M.$$

*Then for a constant $c_0$ depending on $\alpha$, $A$, $L$ and $M$, we have for all $T \geq c_0$,*

$$\mathbb{P}\left(\sup_{g \in \mathcal{G}} \frac{|2(\varepsilon,g)_n|}{\|g\|_n^\alpha} > \frac{T}{\sqrt{n}}\right) \leq c_0 \exp\left(-\frac{T^2}{c_0^2}\right).$$

PROOF OF LEMMA 1. It is clear that $\{g_j/I(g_j)\} = \{g_j : I(g_j) = 1\}$. Hence, by rewriting and then using Lemma 3,

$$\sup_{g_j} \frac{|2(\varepsilon,g_j)_n|}{\|g_j\|_n^\alpha I^{1-\alpha}(g_j)} = \sup_{g_j} \frac{|2(\varepsilon,g_j/I(g_j))_n|}{\|g_j/I(g_j)\|_n^\alpha} \leq \frac{T}{\sqrt{n}}$$

with probability at least $1 - c_0 \exp(-T^2/c_0^2)$. Thus, for $C_0^2 \geq 2c_0^2$ sufficiently large

$$\mathbb{P}\left(\max_j \sup_{g_j} \frac{|2(\varepsilon,g_j)_n|}{\|g_j\|_n^\alpha I^{1-\alpha}(g_j)} > C_0\sqrt{\frac{\log p}{n}}\right)$$
$$\leq pc_0 \exp\left(-\frac{C_0^2 \log p}{c_0^2}\right) \leq c_0 \exp\left(-\frac{C_0^2 \log p}{2c_0^2}\right).$$

In the same spirit, for some constant $c_1$ depending on $L$ and $M$, it holds for all $T \geq c_1$, with probability at least $1 - c_1 \exp(-T^2 d/c_1^2)$,

$$\sup_{h_j} \frac{|2(\varepsilon,h_j)_n|}{\|h_j\|_n} \leq T\sqrt{\frac{d}{n}},$$

where $d$ is the dimension occurring in (6). This result is rather standard but also follows from the more general Corollary 8.3 in [29]. It yields that for $C_1^2 \geq 2c_1^2$, depending on $d$, $L$ and $M$,

$$\max_j \sup_{h_j} \frac{|2(\varepsilon,h_j)_n|}{\|h_j\|_n} \leq C_1 \sqrt{\frac{\log p}{n}}$$

with probability at least $1 - c_1 \exp(-C_1^2 \log p/(2c_1^2))$.

Finally, it is obvious that for all $C_2$ and a constant $c_2$ depending on $L$ and $M$,

$$\mathbb{P}\left(\bar{\varepsilon} > C_2 \sqrt{\frac{\log p}{n}}\right) \leq 2\exp(-C_2^2 \log p / c_2^2).$$

Choosing $c_2 \geq 2$, the result now follows by taking $C = \max\{C_0, C_1, C_2\}$ and $c = c_0 + c_1 + c_2$. $\square$

**Proof of Theorem 1.** We begin with three technical lemmas.
Recall that (for $j = 1, \ldots, p$)

$$\tau_n^2(f_j) = \|f_j\|_n^2 + \lambda^{2-\gamma} I^2(f_j).$$

LEMMA 4. *For $\lambda \geq \sqrt{2}\xi_n/\eta$, we have on $\mathcal{S}_1 \cap \mathcal{S}_2$,*

$$\max_j \sup_{f_j} \frac{2|(\epsilon, f_j)|}{\lambda^{(2-\gamma)/2} \tau_n(f_j)} \leq \eta.$$

PROOF. Note first that with $\lambda \geq \sqrt{2}\xi_n/\eta$,

$$\xi_n \|g_j - g_j^*\|_n^\alpha I^{1-\alpha}(g_j - g_j^*) + \xi_n \|h_j - h_j^*\|_n$$

$$\leq \frac{\eta\lambda}{\sqrt{2}} \|g_j - g_j^*\|_n^\alpha I^{1-\alpha}(g_j - g_j^*) + \frac{\eta\lambda}{\sqrt{2}} \|h_j - h_j^*\|_n$$

$$\leq \eta \frac{\lambda^{(2-\gamma)/2}}{\sqrt{2}} \sqrt{\lambda^{2-\gamma} I^2(g_j - g_j^*) + \|g_j - g_j^*\|_n^2} + \eta \frac{\lambda}{\sqrt{2}} \|h_j - h_j^*\|_n$$

$$\leq \eta \frac{\lambda^{(2-\gamma)/2}}{\sqrt{2}} \sqrt{\lambda^{2-\gamma} I^2(g_j - g_j^*) + \|g_j - g_j^*\|_n^2} + \eta \frac{\lambda^{(2-\gamma)/2}}{\sqrt{2}} \|h_j - h_j^*\|_n,$$

since $\lambda \leq 1$.
We have

$$\sqrt{\lambda^{2-\gamma} I^2(g_j - g_j^*) + \|g_j - g_j^*\|_n^2} + \|h_j - h_j^*\|_n$$

$$\leq \sqrt{2\{\lambda^{2-\gamma} I^2(g_j - g_j^*) + \|g_j - g_j^*\|_n^2 + \|h_j - h_j^*\|_n^2\}}$$

$$= \sqrt{2}\sqrt{\lambda^{2-\gamma} I^2(g_j - g_j^*) + \|f_j - f_j^*\|_n^2},$$

where we used the orthogonality of $g_j - g_j^*$ and $h_j - h_j^*$. The result now follows from the equality $I(g_j - g_j^*) = I(f_j - f_j^*)$. $\square$

It holds that $\hat{c} = \bar{Y}(= \sum_{i=1}^n Y_i/n)$ and $c^* = \mathbb{E}[\bar{Y}]$. Thus, on $\mathcal{S}$, $|\hat{c} - c^*| \leq \xi_n$. Moreover,

$$\|\hat{f} - f_{\text{add}}^0\|_n^2 = |\hat{c} - c^*|^2 + \|(\hat{f} - \hat{c}) - (f_{\text{add}}^0 - c^*)\|_n^2.$$

To simplify the exposition (i.e., avoiding a change of notation), we may therefore assume $\hat{c} = c^*$ and add a $\xi_n^2$ to the final result.

LEMMA 5. *We have on $\mathcal{S}$,*

$$\|\hat{f} - f_{\text{add}}^0\|_n^2 + (1-\eta)\lambda^{(2-\gamma)/2} \sum_{j=1}^p \tau_n(\hat{f}_j - f_j^*) + \lambda^{2-\gamma} \sum_{j=1}^p I^2(\hat{f}_j)$$

$$\leq 2\lambda^{(2-\gamma)/2} \sum_{j \in \mathcal{A}_*} \tau_n(\hat{f}_j - f_j^*) + \lambda^{2-\gamma} \sum_{j \in \mathcal{A}_*} I^2(f_j^*) + \|f^* - f_{\text{add}}^0\|_n^2 + \xi_n^2.$$

PROOF. Because $\hat{f}$ minimizes the penalized loss, we have

$$\frac{1}{n}\sum_{i=1}^n (Y_i - \hat{f}(x_i))^2 + \sum_{j=1}^p J(\hat{f}_j) \leq \frac{1}{n}\sum_{i=1}^n (Y_i - f^*(x_i))^2 + \sum_{j=1}^p J(f_j^*).$$

This can be rewritten as

$$\|\hat{f} - f_{\text{add}}^0\|_n^2 + \sum_{j=1}^p J(\hat{f}_j) \leq 2(\epsilon, \hat{f} - f^*)_n + \sum_{j=1}^p J(f^*) + \|f^* - f_{\text{add}}^0\|_n^2.$$

Thus, on $\mathcal{S}$, by Lemma 4

$$\|\hat{f} - f_{\text{add}}^0\|_n^2 + \sum_{j=1}^p J(\hat{f}_j) \leq \eta\lambda^{(2-\gamma)/2} \sum_{j=1}^p \tau_n(\hat{f}_j - f_j^*) + \sum_{j=1}^p J(f_j^*)$$

$$+ \|f^* - f_{\text{add}}^0\|_n^2$$

or

$$\|\hat{f} - f_{\text{add}}^0\|_n^2 + \sum_{j \notin \mathcal{A}_*} \lambda^{(2-\gamma)/2} \tau_n(\hat{f}_j) + \lambda^{2-\gamma} \sum_{j=1}^p I^2(\hat{f}_j)$$

$$\leq \eta\lambda^{(2-\gamma)/2} \sum_{j=1}^p \tau_n(\hat{f}_j - f_j^*) + \lambda^{(2-\gamma)/2} \sum_{j \in \mathcal{A}_*} (\tau_n(f_j^*) - \tau_n(\hat{f}_j))$$

$$+ \lambda^{2-\gamma} \sum_{j \in \mathcal{A}_*} I^2(f_j^*) + \|f^* - f_{\text{add}}^0\|_n^2$$

$$\leq (1+\eta)\lambda^{(2-\gamma)/2} \sum_{j \in \mathcal{A}_*} \tau_n(\hat{f}_j - f_j^*) + \eta\lambda^{(2-\gamma)/2} \sum_{j \notin \mathcal{A}_*} \tau_n(\hat{f}_j - f_j^*)$$

$$+ \lambda^{2-\gamma} \sum_{j \in \mathcal{A}_*} I^2(f_j^*) + \|f^* - f_{\text{add}}^0\|_n^2.$$

In other words,

$$\|\hat{f} - f_{\text{add}}^0\|_n^2 + (1-\eta)\lambda^{(2-\gamma)/2} \sum_{j \notin \mathcal{A}_*} \tau_n(\hat{f}_j) + \lambda^{2-\gamma} \sum_{j=1}^{p} I^2(\hat{f}_j)$$
$$\leq (1+\eta)\lambda^{(2-\gamma)/2} \sum_{j \in \mathcal{A}_*} \tau_n(\hat{f}_j - f_j^*) + \lambda^{2-\gamma} \sum_{j \in \mathcal{A}_*} I^2(f_j^*) + \|f^* - f_{\text{add}}^0\|_n^2,$$

so that

$$\|\hat{f} - f_{\text{add}}^0\|_n^2 + (1-\eta)\sum_{j=1}^{p} \lambda^{(2-\gamma)/2}\tau_n(\hat{f}_j - f_j^*) + \lambda^{2-\gamma} \sum_{j=1}^{p} I^2(\hat{f}_j)$$
$$\leq 2\lambda^{(2-\gamma)/2} \sum_{j \in \mathcal{A}_*} \tau_n(\hat{f}_j - f_j^*) + \lambda^{2-\gamma} \sum_{j \in \mathcal{A}_*} I^2(f_j^*) + \|f^* - f_{\text{add}}^0\|_n^2. \quad \square$$

COROLLARY 3. *On $\mathcal{S}$, either*

(12)
$$\|\hat{f} - f^*\|_n^2 + (1-\eta)\lambda^{(2-\gamma)/2} \sum_{j=1}^{p} \tau_n(\hat{f}_j - f_j^*) + \lambda^{2-\gamma} \sum_{j=1}^{p} I^2(\hat{f}_j)$$
$$\leq 4\lambda^{(2-\gamma)/2} \sum_{j \in \mathcal{A}_*} \tau_n(\hat{f}_j - f_j^*)$$

*or*

(13)
$$\|\hat{f} - f^*\|_n^2 + (1-\eta)\lambda^{(2-\gamma)/2} \sum_{j=1}^{p} \tau_n(\hat{f}_j - f_j^*) + \lambda^{2-\gamma} \sum_{j=1}^{p} I^2(\hat{f}_j)$$
$$\leq 2\lambda^{2-\gamma} \sum_{j \in \mathcal{A}_*} I^2(f_j^*) + 2\|f^* - f_{\text{add}}^0\|_n^2 + 2\xi_n^2.$$

Observe that if (13) holds, we have nothing further to prove, as this is already an oracle inequality. So we only have to work with (12). It implies that

(14) $$\sum_{j=1}^{p} \tau_n(\hat{f}_j - f_j^*) \leq \frac{4}{1-\eta} \sum_{j \in \mathcal{A}_*} \tau_n(\hat{f}_j - f_j^*),$$

in other words, we may apply the compatibility condition to $\hat{f} - f^*$.

LEMMA 6. *Suppose the compatibility condition holds. Then (14) implies*

$$4\lambda^{(2-\gamma)/2} \sum_{j \in \mathcal{A}_*} \tau_n(\hat{f}_j - f_j^*) \leq 24 \frac{\lambda^{2-\gamma}|\mathcal{A}_*|}{\phi_{n,*}^2} + \lambda^{2-\gamma} \sum_{j \in \mathcal{A}_*} (I^2(\hat{f}_j) + I^2(f_j^*))$$
$$+ \|\hat{f} - f_{\text{add}}^0\|_n^2 + \|f^* - f_{\text{add}}^0\|_n^2$$

*(under the simplifying assumption $\hat{c} = c^* = 0$).*

PROOF. We have
$$4\lambda^{(2-\gamma)/2} \sum_{j\in\mathcal{A}_*} \tau_n(\hat{f}_j - f_j^*)$$
$$\leq 4\lambda^{(2-\gamma)/2}\sqrt{|\mathcal{A}_*|}\sqrt{\sum_{j\in\mathcal{A}_*} \|\hat{f}_j - f_j^*\|_n^2 + \lambda^{2-\gamma}\sum_{j\in\mathcal{A}_*} I^2(\hat{f}_j - f_j^*)}.$$

The compatibility condition now gives
$$4\lambda^{(2-\gamma)/2} \sum_{j\in\mathcal{A}_*} \tau_n(\hat{f}_j - f_j^*)$$
$$\leq \frac{4\lambda^{(2-\gamma)/2}\sqrt{|\mathcal{A}_*|}}{\phi_{n,*}}\sqrt{\left\|\sum_{j=1}^p (\hat{f}_j - f_j^*)\right\|_n^2 + 2\lambda^{2-\gamma}\sum_{j\in\mathcal{A}_*} I^2(\hat{f}_j - f_j^*)}.$$

With the simplifying assumption $\hat{c} = c^* = 0$, we may use the shorthand notation $\hat{f} = \sum_j \hat{f}_j$ and $f^* = \sum_j f_j^*$. Next, we apply the triangle inequality:
$$\sqrt{\|\hat{f} - f^*\|_n^2 + 2\lambda^{2-\gamma}\sum_{j\in\mathcal{A}_*} I^2(\hat{f}_j - f_j^*)}$$
$$\leq \|\hat{f} - f_{\text{add}}^0\|_n + \|f^* - f_{\text{add}}^0\|_n$$
$$\quad + \sqrt{2\lambda^{2-\gamma}\sum_{j\in\mathcal{A}_*} I^2(\hat{f}_j)} + \sqrt{2\lambda^{2-\gamma}\sum_{j\in\mathcal{A}_*} I^2(f_j^*)}.$$

We now use
$$\frac{4\lambda^{(2-\gamma)/2}\sqrt{|\mathcal{A}_*|}}{\phi_{n,*}}\|\hat{f} - f_{\text{add}}^0\|_n \leq \frac{4\lambda^{2-\gamma}|\mathcal{A}_*|}{\phi_{n,*}^2} + \|\hat{f} - f_{\text{add}}^0\|_n^2$$

and similarly with $\hat{f}$ replaced by $f^*$. In the same spirit
$$\frac{4\lambda^{(2-\gamma)/2}\sqrt{|\mathcal{A}_*|}}{\phi_{n,*}}\sqrt{2\lambda^{2-\gamma}\sum_{j\in\mathcal{A}_*} I^2(\hat{f}_j)}$$
$$\leq \frac{8\lambda^{2-\gamma}|\mathcal{A}_*|}{\phi_{n,*}^2} + \lambda^{2-\gamma}\sum_{j\in\mathcal{A}_*} I^2(\hat{f}_j)$$

and similarly with $\hat{f}$ replaced by $f^*$. □

PROOF OF THEOREM 1. By Lemma 5, we have on $\mathcal{S}$,
$$\|\hat{f} - f_{\text{add}}^0\|_n^2 + (1-\eta)\lambda^{(2-\gamma)/2}\sum_{j=1}^p \tau_n(\hat{f}_j - f_j^*) + \lambda^{2-\gamma}\sum_{j=1}^p I^2(\hat{f}_j)$$

$$\leq 2\lambda^{(2-\gamma)/2} \sum_{j \in \mathcal{A}_*} \tau_n(\hat{f}_j - f_j^*) + \lambda^{2-\gamma} \sum_{j \in \mathcal{A}_*} I^2(f_j^*)$$
$$+ \|f^* - f_{\text{add}}^0\|_n^2 + \xi_n^2.$$

In view of Corollary 3, we can assume without loss of generality that (12) holds. Lemma 6 tells us now that

$$\|\hat{f} - f_{\text{add}}^0\|_n^2 + (1-\eta)\lambda^{(2-\gamma)/2} \sum_{j=1}^p \tau_n(\hat{f}_j - f_j^*) + \lambda^{2-\gamma} \sum_{j=1}^p I^2(\hat{f}_j)$$
$$\leq 12\frac{\lambda^{2-\gamma}|\mathcal{A}_*|}{\phi_{n,*}^2} + \frac{1}{2}\lambda^{2-\gamma} \sum_{j \in \mathcal{A}_*} I^2(\hat{f}_j) + \frac{1}{2}\|\hat{f} - f_{\text{add}}^0\|_n^2 + \frac{3}{2}\|f^* - f_{\text{add}}^0\|_n^2$$
$$+ \frac{3}{2}\lambda^{2-\gamma} \sum_{j \in \mathcal{A}_*} I^2(f_j^*) + \xi_n^2.$$

This can be rewritten as

$$\|\hat{f} - f_{\text{add}}^0\|_n^2 + 2(1-\eta)\lambda^{(2-\gamma)/2} \sum_{j=1}^p \tau_n(\hat{f}_j - f_j^*) + \lambda^{2-\gamma} \sum_{j=1}^p I^2(\hat{f}_j)$$
$$\leq 24\frac{\lambda^{2-\gamma}|\mathcal{A}_*|}{\phi_{n,*}^2} + 3\|f^* - f_{\text{add}}^0\|_n^2 + 3\lambda^{2-\gamma} \sum_{j \in \mathcal{A}_*} I^2(f_j^*) + 2\xi_n^2. \quad \square$$

**A.1. Proof of Theorem 2.** We first show that the $\|\cdot\|$-norm and the $\|\cdot\|_n$-norm are in some sense compatible, and then prove the same for the norms $\tau$ and $\tau_n$.

LEMMA 7. *Suppose the theoretical compatibility condition holds, and that*
$$\frac{2C_0 c_\eta^2 |\mathcal{A}_*|\lambda^{1-\gamma}}{\phi_*^2} \leq 1.$$
*Then on* $\mathcal{S}_4$, *for all* $f$ *satisfying*
$$\tau_{\text{tot}}(f) \leq c_\eta \tau_{\text{in}}(f),$$
*we have*
$$\|f\|^2 \leq 2\|f\|_n^2 + (1 + \phi_*^2) \sum_{j \in \mathcal{A}_*} \lambda^{2-\gamma} I^2(f_j).$$

PROOF.
$$\|f\|^2 \leq \|f\|_n^2 + C_0 \lambda^{1-\gamma} \tau_{\text{tot}}^2(f)$$
$$\leq \|f\|_n^2 + C_0 c_\eta^2 \lambda^{1-\gamma} \tau_{\text{in}}^2(f)$$

$$\leq \|f\|_n^2 + C_0 c_\eta^2 \lambda^{1-\gamma} |\mathcal{A}_*| \sum_{j \in \mathcal{A}_*} (\|f_j\|^2 + \lambda^{2-\gamma} I^2(f_j))$$

$$\leq \|f\|_n^2 + \frac{\phi_*^2}{2} \sum_{j \in \mathcal{A}_*} (\|f_j\|^2 + \lambda^{2-\gamma} I^2(f_j))$$

$$\leq \|f\|_n^2 + \frac{1}{2} \|f\|^2 + \frac{1+\phi_*^2}{2} \sum_{j \in \mathcal{A}_*} \lambda^{2-\gamma} I^2(f_j). \qquad \Box$$

LEMMA 8. *On the set* $\mathcal{S}_4$, *and for* $\lambda^{1-\gamma} C_0 < 1$, *it holds that*
$$(1 - \lambda^{1-\gamma} C_0) \tau(f_j) \leq \tau_n(f_j) \leq (1 + \lambda^{1-\gamma} C_0) \tau(f_j)$$
*for all* $j$.

PROOF.
$$|\tau_n(f_j) - \tau(f_j)| \leq \frac{|\|f_j\|_n^2 - \|f_j\|^2|}{\tau(f_j)}$$
$$\leq \frac{\lambda^{1-\gamma} C_0 \tau^2(f_j)}{\tau(f_j)}. \qquad \Box$$

We use the short-hand notation
$$\hat{\tau}_{\text{in}}(f) = \sum_{j \in \mathcal{A}_*} \tau_n(f_j), \qquad \hat{\tau}_{\text{out}}(f) = \sum_{j \notin \mathcal{A}_*} \tau_n(f_j)$$
and
$$\hat{\tau}_{\text{tot}}(f) = \hat{\tau}_{\text{in}}(f) + \hat{\tau}_{\text{out}}(f).$$

PROOF OF THEOREM 2. If
$$\hat{\tau}_{\text{tot}}(f) \leq \frac{4}{1-\eta} \hat{\tau}_{\text{in}}(f),$$
then by Lemma 8, on $\mathcal{S}_4$,
$$\tau_{\text{tot}}(f) \leq \frac{4(1+\eta)}{(1-\eta)^2} \tau_{\text{in}}(f).$$
Moreover, on $\mathcal{S}_4$, for all $j$
$$\|f_j\|_n^2 \leq \|f_j\|^2 + \eta \tau^2(f_j).$$
Hence,
$$\sum_{j \in \mathcal{A}_*} \|f_j\|_n^2 \leq \sum_{j \in \mathcal{A}_*} \|f_j\|^2 + \eta \tau_{\text{in}}^2(f)$$
$$= (1+\eta) \sum_{j \in \mathcal{A}_*} \|f_j\|^2 + \eta \lambda^{2-\gamma} \sum_{j \in \mathcal{A}_*} I^2(f_j).$$

Applying the theoretical compatibility condition, we arrive at

$$\sum_{j\in\mathcal{A}_*} \|f_j\|_n^2 \leq \frac{(1+\eta)}{\phi_*^2}\left(\|f\|^2 + \lambda^{2-\gamma}\sum_{j\in\mathcal{A}_*} I^2(f_j)\right) + \eta\lambda^{2-\gamma}\sum_{j\in\mathcal{A}_*} I^2(f_j)$$

$$= \frac{(1+\eta)}{\phi_*^2}\|f\|^2 + \left(\frac{(1+\eta)}{\phi_*^2} + \eta\right)\lambda^{2-\gamma}\sum_{j\in\mathcal{A}_*} I^2(f_j).$$

Next, apply Lemma 7 to obtain

$$\sum_{j\in\mathcal{A}_*} \|f_j\|_n^2 \leq \frac{2(1+\eta)}{\phi_*^2}\|f\|_n^2$$

$$+ \left((1+\eta)(1+\phi_*^2) + \frac{(1+\eta)}{\phi_*^2} + \eta\right)\lambda^{2-\gamma}\sum_{j\in\mathcal{A}_*} I^2(f_j)$$

$$\leq \left((1+\eta)(1+\phi_*^2) + \frac{2(1+\eta)}{\phi_*^2} + \eta\right)$$

$$\times \left(\|f\|_n^2 + \lambda^{2-\gamma}\sum_{j\in\mathcal{A}_*} I^2(f_j)\right). \qquad \square$$

## APPENDIX B: THE SET $\mathcal{S}_4$

In this subsection, we show that the set $\mathcal{S}_4$ has large probability, under reasonable conditions (mainly Condition D below). We assume again throughout that the functions $f_j$ are centered with respect to the theoretical measure $Q$. (Our estimator of course uses the empirical centering. It is not difficult to see that this difference can be taken care of by adding a term of order $1/\sqrt{n}$ in the oracle result.)

Let $\mu$ be Lebesgue measure on $[0,1]$, and let for $f_j: [0,1] \to \mathbb{R}$,

$$I^2(f_j) = \int |f_j^{(s)}|^2 \, d\mu = \|f_j^{(s)}\|_\mu^2,$$

where $\|\cdot\|_\mu$ denotes the $L_2(\mu)$-norm. Moreover, write $\mathcal{F}_j = \{f_j : I(f_j) < \infty\}$. We let

$$\alpha = 1 - \frac{1}{2s}$$

and

$$\gamma = \frac{2(1-\alpha)}{2-\alpha}$$

as before.

We will use symmetrization arguments, and therefore introduce a Rademacher sequence $\{\sigma_i\}$, independent of $\{X_i\}$.

The argumentation we shall employ can be summarized as follows. By a contraction argument, we make the transition from the $f^2$'s to the $f$'s. This step needs boundedness of weighted $f$'s, because the function $x \mapsto x^2$ is only Lipschitz on a bounded interval. The fact that we use the Sobolev norm as measure of complexity makes this work. The contraction inequality is in terms of the *expectation* of the weighted empirical process. We use a concentration inequality to get a hold on the *probabilities*.

The original $f$'s are handled by looking at the maximum over $j$ of the weighted empirical process indexed by $\mathcal{F}_j$. This is done by first bounding the expectation, then applying a concentration inequality to get exponentially small probabilities. This allows us to get similar probability inequalities uniformly in $j \in \{1,\ldots,p\}$, inserting a $\log p$-term. We then rephrase the probabilities back to expectations, now uniformly in $j$.

To establish a bound for the expectation of the weighted empirical process indexed by $\mathcal{F}_j$ with $j$ fixed, we first prove a conditional bound involving the empirical norm, then a contraction inequality to reduce the problem of this empirical norm, involving the $f_j^2$'s, to the problem involving the original $f_j$'s. We then unravel the knot.

We now will present this program, but in reverse order.

**B.1. Weighted empirical process for fixed $j$.** We fix an arbitrary $j \in \{1,\ldots,p\}$, and consider the weighted empirical process

$$\frac{|1/n \sum_{i=1}^n \sigma_i f_j(X_i^{(j)})|}{\lambda^{(2-\gamma)/2}\tau(f_j)}.$$

Our aim is to prove Corollary 5.

The following lemma is well known in the approximation literature. We refer to [29] and the references therein. For a class of functions $\mathcal{G}$, we denote the entropy of $\mathcal{G}$ endowed with the metric induced by the sup-norm, by $H_\infty(\cdot, \mathcal{G})$.

LEMMA 9. *For some constant $A_s$, we have*

$$H_\infty(\delta, \{I(f_j) \leq 1, |f_j|_\infty \leq 1\}) \leq \frac{A_s^2}{\alpha^2}\delta^{-2(1-\alpha)}, \qquad \delta > 0.$$

Let for all $R > 0$,

$$\mathcal{F}_j(R) = \{I(f_j) \leq 1, |f_j|_\infty \leq 1, \|f_j\| \leq R\}.$$

The next theorem is along the lines of, for example, [31], Corollary 2.2.5. It applies the entropy bound of Lemma 9. We have put in a rough but explicit constant. We write $\mathbb{E}_\mathbf{X}$ for the conditional expectation given $\mathbf{X} = (X_1, \ldots, X_n)$.

THEOREM 3. *We have*
$$\mathbb{E}_{\mathbf{X}}\left[\sup_{f_j \in \mathcal{F}_j(R)} \left|\frac{1}{n}\sum_{i=1}^n \sigma_i f_j(X_i^{(j)})\right|\right] \leq \frac{16 A_s}{\sqrt{n}} \hat{R}_n^\alpha,$$

*where*
$$\hat{R}_n = \sup_{f_j \in \mathcal{F}_j(R)} \|f_j\|_n.$$

To turn the bound of Theorem 3 into a bound for the unconditional expectation, we need to handle the random $\hat{R}_n$. For this purpose, we reuse Theorem 3 itself.

THEOREM 4. *We have*
$$\mathbb{E}[\hat{R}_n^\alpha] \leq \sqrt{(2R^2)^\alpha + \left(2^8 \frac{A_s}{\sqrt{n}}\right)^{2\alpha(2-\gamma)/2}}.$$

PROOF. By symmetrization and the contraction inequality of [18],
$$\mathbb{E}\left[\sup_{f_j \in \mathcal{F}_j(R)} |\|f_j\|_n^2 - \|f_j\|^2|\right] \leq 8\mathbb{E}\left[\sup_{f_j \in \mathcal{F}_j(R)} \left|\frac{1}{n}\sum_{i=1}^n \sigma_i f_j(X_i^{(j)})\right|\right]$$
$$\leq 2^7 \frac{A_s}{\sqrt{n}} \mathbb{E}[\hat{R}_n^\alpha],$$

where we used Theorem 3. It also follows that
$$\mathbb{E}[\hat{R}_n^2] - R^2 \leq 2^7 \frac{A_s}{\sqrt{n}} \mathbb{E}[\hat{R}_n^\alpha].$$

Since by Jensen's inequality
$$\mathbb{E}[\hat{R}_n^2] = \mathbb{E}[(\hat{R}_n^\alpha)^{2/\alpha}] \geq (\mathbb{E}[\hat{R}_n^\alpha])^{2/\alpha},$$

we may conclude that
$$(\mathbb{E}[\hat{R}_n^\alpha])^{2/\alpha} \leq R^2 + 2^7 \frac{A_s}{\sqrt{n}} \mathbb{E}[\hat{R}_n^\alpha].$$

Now, for any positive $a$ and $b$,
$$ab \leq a^{2/(2-\alpha)} + b^{2/\alpha},$$

hence, also
$$ab \leq 2^{\alpha/(2-\alpha)} a^{2/(2-\alpha)} + \tfrac{1}{2} b^{2/\alpha}.$$

Apply this with
$$a = 2^7 \frac{A_s}{\sqrt{n}}, \qquad b = \mathbb{E}[\hat{R}_n^\alpha],$$

to find

$$(\mathbb{E}[\hat{R}_n^\alpha])^{2/\alpha} \leq R^2 + 2^{\alpha/(2-\alpha)} \left(2^7 \frac{A_s}{\sqrt{n}}\right)^{2/(2-\alpha)} + \frac{1}{2}(\mathbb{E}[\hat{R}_n^\alpha])^{2/\alpha}.$$

It follows that

$$(\mathbb{E}[\hat{R}_n^\alpha])^{2/\alpha} \leq 2R^2 + \left(2^8 \frac{A_s}{\sqrt{n}}\right)^{2/(2-\alpha)}$$

and hence

$$\mathbb{E}[\hat{R}_n^\alpha] \leq \sqrt{2^\alpha R^{2\alpha} + \left(2^8 \frac{A_s}{\sqrt{n}}\right)^{2\alpha/(2-\alpha)}}$$

$$= \sqrt{(2R^2)^\alpha + \left(2^8 \frac{A_s}{\sqrt{n}}\right)^{2\alpha(2-\gamma)/2}}. \qquad \square$$

COROLLARY 4. *We have*

$$\mathbb{E}\left[\sup_{f_j \in \mathcal{F}_j(R)} \left|\frac{1}{n}\sum_{i=1}^n \sigma_i f_j(X_i^{(j)})\right|\right] \leq \frac{2^4 A_s}{\sqrt{n}} \sqrt{(2R^2)^\alpha + \left(2^8 \frac{A_s}{\sqrt{n}}\right)^{2\alpha(2-\gamma)/2}}$$

$$\leq \frac{\tilde{A}_s}{\sqrt{n}} \sqrt{R^{2\alpha} + \left(\frac{\tilde{A}_s}{\sqrt{n}}\right)^{(2-\gamma)}}$$

*for some constant $\tilde{A}_s$ depending only on $\alpha = \alpha(s)$ and $A_s$.*

The peeling device is inserted to establish a bound for the weighted empirical process.

LEMMA 10. *Define*

$$\delta_n = (\tilde{A}_s/\sqrt{n}).$$

*Then for $\lambda \geq \delta_n$,*

$$\mathbb{E}\left[\sup_{I(f_j)\leq 1, |f_j|_\infty \leq 1} \frac{|1/n \sum_{i=1}^n \sigma_i f_j(X_i^{(j)})|}{\lambda^{(2-\gamma)/2}\sqrt{\|f_j\|^2 + \lambda^{2-\gamma}}}\right] \leq C_s \frac{\delta_n}{\lambda},$$

*where*

$$C_s = 2\left(1 + \frac{\alpha^{-\alpha/(1-\alpha)}}{1-\alpha}\right).$$

PROOF. Set $z = \alpha^{-1/(1-\alpha)}$. Then

$$\mathbb{E}\left[\sup_{I(f_j)\leq 1, |f_j|_\infty \leq 1} \frac{|1/n \sum_{i=1}^n \sigma_i f_j(X_i^{(j)})|}{\lambda^{(2-\gamma)/2}\sqrt{\|f_j\|^2 + \lambda^{2-\gamma}}}\right]$$

$$\leq \mathbb{E}\left[\sup_{f_j \in \mathcal{F}_j(\lambda^{(2-\gamma)/2})} \frac{|1/n \sum_{i=1}^n \sigma_i f_j(X_i^{(j)})|}{\lambda^{2-\gamma}}\right]$$

$$+ \sum_{j=1}^\infty \mathbb{E}\left[\sup_{f_j \in \mathcal{F}_j(z^j \lambda^{(2-\gamma)/2})} \frac{|1/n \sum_{i=1}^n \sigma_i f_j(X_i^{(j)})|}{\lambda^{2-\gamma} z^{j-1}}\right]$$

$$\leq \frac{2\delta_n \lambda^{1-\gamma}}{\lambda^{2-\gamma}} + \sum_{j=1}^\infty \frac{2\delta_n \lambda^{1-\gamma} z^{j\alpha}}{\lambda^{2-\gamma} z^{j-1}} \leq 2 + \sum_{j=1}^\infty \frac{\delta_n^2 z^{j\alpha} + \delta_n^2}{\delta_n^2 z^{j-1}}$$

$$\leq \left(2 + 2z \sum_{j=1}^\infty z^{-j(1-\alpha)}\right)\frac{\delta_n}{\lambda} = \left(2 + 2\frac{\alpha^{-\alpha/(1-\alpha)}}{1-\alpha}\right)\frac{\delta_n}{\lambda}. \quad \square$$

We now show how to get rid of the restriction $|f_j|_\infty \leq 1$ in Lemma 10.

LEMMA 11. *Define*

$$\delta_n = \tilde{A}_s/\sqrt{n}.$$

*Then for $\delta_n \leq \lambda \leq 1$,*

$$\mathbb{E}\left[\sup_{I(f_j)\leq 1} \frac{|1/n \sum_{i=1}^n \sigma_i f_j(X_i^{(j)})|}{\lambda^{(2-\gamma)/2}\sqrt{\|f_j\|^2 + \lambda^{2-\gamma}}}\right] \leq \tilde{C}_s \frac{\delta_n}{\lambda},$$

*where*

$$\tilde{C}_s = \sqrt{s-1} + C_s.$$

PROOF. We can write $f_j = g_j + h_j$, where $h_j$ is a polynomial of degree $s-1$ and $|g_j|_\infty \leq I(g_j) = I(f_j)$. We take $g_j$ and $h_j$ are orthogonal:

$$\int g_j h_j \, dQ_j = 0.$$

Then

$$\frac{|1/n \sum_{i=1}^n \sigma_i f_j(X_i^{(j)})|}{\lambda^{(2-\gamma)/2}\sqrt{\|f_j\|^2 + \lambda^{2-\gamma}}} \leq \frac{|1/n \sum_{i=1}^n \sigma_i g_j(X_i^{(j)})|}{\lambda^{(2-\gamma)/2}\sqrt{\|g_j\|^2 + \lambda^{2-\gamma}}} + \frac{|1/n \sum_{i=1}^n \sigma_i h_j(X_i^{(j)})|}{\lambda^{(2-\gamma)/2}\|h_j\|}.$$

We moreover can write

$$h_j = \sum_{k=1}^{s-1} \theta_k p_k(\cdot),$$

where the $\{p_k\}$ are orthogonal polynomials, and have norm $\|p_k\| = 1$. Hence, using that $\sum_{k=1}^{s-1} \theta_k^2 = \|h_j\|^2$,

$$\frac{|1/n \sum_{i=1}^n \sigma_i h_j(X_i^{(j)})|}{\|h_j\|} \leq \sqrt{\sum_{k=1}^{s-1} \left(\frac{1}{n}\left|\sum_{i=1}^n \sigma_i p_k(X_i^{(j)})\right|\right)^2}.$$

This gives

$$\mathbb{E}\left[\sup_{h_j} \frac{|1/n \sum_{i=1}^n \sigma_i h_j(X_i^{(j)})|}{\lambda^{(2-\gamma)/2}\|h_j\|}\right] \leq \frac{\sqrt{s-1}}{\lambda^{(2-\gamma)/2}\sqrt{n}} \leq \sqrt{s-1}\frac{\delta_n}{\lambda},$$

since

$$\sqrt{n}\delta_n = \tilde{A}_s \geq 1. \qquad \square$$

Using the renormalization

$$f_j \mapsto f_j/I(f_j)$$

we arrive at the required result:

COROLLARY 5. *We have*

$$\mathbb{E}\left[\sup_{f_j} \frac{|1/n \sum_{i=1}^n \sigma_i f_j(X_i^{(j)})|}{\lambda^{(2-\gamma)/2}\sqrt{\|f_j\|^2 + \lambda^{2-\gamma}I^2(f_j)}}\right] \leq \tilde{C}_s \frac{\delta_n}{\lambda}.$$

**B.2. From expectation to probability and back.** Let $\mathcal{G}$ be some class of functions on $\mathcal{X}$, $\zeta_1, \ldots, \zeta_n$ be independent random variables with values in $\mathcal{X}$, and

$$Z = \sup_{g \in \mathcal{G}} \left|\frac{1}{n}\sum_{i=1}^n (g(\zeta_i) - \mathbb{E}[g(\zeta_i)])\right|.$$

Concentration inequalities are exponential probability inequalities for the amount of concentration of $Z$ around its mean. We present here a very tight concentration inequality, which was established by [4].

THEOREM 5 (Bousquet's concentration theorem [4]). *Suppose*

$$\frac{1}{n}\sum_{i=1}^n \mathbb{E}[(g(\zeta_i) - \mathbb{E}[g(\zeta_i)])^2] \leq R^2 \qquad \forall g \in \mathcal{G},$$

*and moreover, for some positive constant* $K$,

$$|g(\zeta_i) - \mathbb{E}[g(\zeta_i)]| \leq K \qquad \forall g \in \mathcal{G}.$$

*We have for all* $t > 0$,

$$\mathbb{P}\left(Z \geq \mathbb{E}[Z] + \frac{tK}{3n} + \sqrt{\frac{2t(R^2 + 2K\mathbb{E}[Z])}{n}}\right) \leq \exp(-t).$$

COROLLARY 6. *Under the conditions of Theorem 5,*

(15) $$\mathbb{P}\left(Z \geq 4\mathbb{E}[Z] + \frac{2tK}{3n} + R\sqrt{\frac{2t}{n}}\right) \leq \exp(-t).$$

Converse, given an exponential probability inequality, one can of course prove an inequality for the expectation.

LEMMA 12. *Let $Z \geq 0$ be a random variable, satisfying for some constants $C_1$, $L$ and $M$,*

$$\mathbb{P}\left(Z \geq C_1 + \frac{Lt}{n} + M\sqrt{\frac{2t}{n}}\right) \leq \exp(-t) \qquad \forall t > 0.$$

*Then*

$$\mathbb{E}[Z] \leq C_1 + \frac{L}{n} + M\sqrt{\frac{\pi}{2n}}.$$

PROOF.
$$\mathbb{E}[Z] = \int_0^\infty \mathbb{P}(Z \geq a)\,da \leq C_1 + \int_0^\infty \mathbb{P}(Z > C_1 + a)\,da.$$

Now, use the change of variables

$$a = \frac{Lt}{n} + M\sqrt{\frac{2t}{n}}.$$

Then

$$da = \left(\frac{L}{n} + \frac{M}{\sqrt{2nt}}\right)dt.$$

So

$$\mathbb{E}[Z] \leq C_1 + \frac{L}{n}\int_0^\infty e^{-t}\,dt + \frac{M}{\sqrt{2n}}\int_0^\infty e^{-t}/\sqrt{t}\,dt$$

$$= C_1 + \frac{L}{n} + M\sqrt{\frac{\pi}{2n}}. \qquad \square$$

LEMMA 13. *Let, for $j = 1, \ldots, p$, $\mathcal{G}_j$ be a class of functions and let*

$$Z_j = \sup_{g_j \in \mathcal{G}_j} \left|\frac{1}{n}\sum_{i=1}^n \sigma_i g_j(X_i)\right|.$$

*Suppose that for all $j$ and all $g_j \in \mathcal{G}_j$,*

$$\|g_j\| \leq R, \qquad |g_j|_\infty \leq K.$$

*Then*

$$\mathbb{E}\Big[\max_{1\leq j\leq p} Z_j\Big] \leq 4 \max_{1\leq j\leq p} \mathbb{E}[Z_j] + \frac{2K(1+\log p)}{3n} + R\sqrt{\frac{4(1+\log p)}{n}}.$$

PROOF. Let

$$E_j = \mathbb{E}[Z_j].$$

Then by the corollary of Bousquet's inequality, we have

$$\mathbb{P}\bigg(Z_j \geq 4E_j + \frac{2Kt}{3n} + R\sqrt{\frac{2t}{n}}\bigg) \leq \exp(-t) \qquad \forall t > 0.$$

Replacing $t$ by $t + \log p$, one finds that

$$\mathbb{P}\bigg(\max_j Z_j \geq 4\max_j E_j + \frac{2Kt}{3n} + \frac{4K\log p}{3n} + R\sqrt{\frac{2t}{n}} + R\sqrt{\frac{2\log p}{n}}\bigg)$$
$$\leq p\exp[-(t+\log p)] = \exp(-t).$$

Apply Lemma 12, with the bound $\pi/4 \leq 1$, and with

$$C_1 = 4\max_j E_j + \frac{2K\log p}{3n} + R\sqrt{\frac{2\log p}{n}},$$
$$L = \frac{2K}{3}, \qquad M = R. \qquad \square$$

**B.3. The supremum norm.** The following lemma can be found in [29]. It is a corollary of the interpolation inequality of [1].

LEMMA 14. *There exists a constant $c_s$ such that for all $f_j$ with $I(f_j) \leq 1$, one has*

$$|f_j|_\infty \leq c_s \|f_j\|_\mu^\alpha.$$

CONDITION D. For all $j$, $dQ_j/d\mu = q_j$ exists and

$$q_j \geq \eta_0^2 > 0.$$

COROLLARY 7. *Assume Condition D. Then for all $j$ and all $f_j$ with $I(f_j) \leq 1$, we have*

$$|f_j|_\infty \leq c_{s,q}\|f_j\|^\alpha,$$

*where $c_{s,q} = c_s/\eta_0$. This implies that for all $j$ and $f_j$,*

$$|f_j|_\infty \leq c_{s,q}\|f_j\|^\alpha I^{1-\alpha}(f_j).$$

**B.4. Expectation uniformly over $j \in \{1, \ldots, p\}$.**

LEMMA 15. *Assume Condition D and that $\lambda \geq \sqrt{4(1+\log p)/n}$, and $\delta_n \leq \lambda \leq 1$. We have*

$$\mathbb{E}\left[\max_j \sup_{f_j} \frac{|1/n \sum_{i=1}^n \sigma_i f_j(X_i^{(j)})|}{\lambda^{(2-\gamma)/2} \tau(f_j)}\right]$$

$$\leq 4\tilde{C}_s \frac{\delta_n}{\lambda} + c_{s,q} \lambda + \lambda^{\gamma/2}.$$

PROOF. By Corollary 5, we have for each $j$

$$\mathbb{E}\left[\sup_{f_j} \frac{|1/n \sum_{i=1}^n \sigma_i f_j(X_i^{(j)})|}{\lambda^{(2-\gamma)/2} \tau(f_j)}\right] \leq \tilde{C}_s \frac{\delta_n}{\lambda}.$$

Moreover, in view of Corollary 7,

$$\frac{|f_j|_\infty}{\lambda^{(2-\gamma)/2} \tau(f_j)} \leq \frac{c_{s,q}}{\lambda}.$$

We also have

$$\frac{\|f_j\|}{\lambda^{(2-\gamma)/2} \tau(f_j)} \leq \frac{1}{\lambda^{(2-\gamma)/2}}.$$

Now, apply Lemma 13 with

$$K = \frac{c_{s,q}}{\lambda}, \qquad R = \frac{1}{\lambda^{(2-\gamma)/2}},$$

to find

$$\mathbb{E}\left[\max_j \sup_{f_j} \frac{|1/n \sum_{i=1}^n \sigma_i f_j(X_i^{(j)})|}{\lambda^{(2-\gamma)/2} \tau(f_j)}\right]$$

$$\leq 4\tilde{C}_s \frac{\delta_n}{\lambda} + \frac{2c_{s,q}(1+\log p)}{3n\lambda} + \frac{1}{\lambda^{(2-\gamma)/2}} \sqrt{\frac{4(1+\log p)}{n}}. \qquad \square$$

**B.5. Expectation of the weighted empirical process, indexed by the additive $f$'s.**

LEMMA 16. *Assume Condition D and that $\lambda \geq \sqrt{4(1+\log p)/n}$, and $\delta_n \leq \lambda \leq 1$. Then*

$$\mathbb{E}\left[\sup_f \frac{|1/n \sum_{i=1}^n \sigma_i f(X_i)|}{\lambda^{(2-\gamma)/2} \tau_{\text{tot}}(f)}\right]$$

$$\leq 4\tilde{C}_s \frac{\delta_n}{\lambda} + c_{s,q} \lambda + \lambda^{\gamma/2}.$$

Proof. It holds that
$$\left|\frac{1}{n}\sum_{i=1}^n \sigma_i f(X_i)\right| \le \sum_{j=1}^p \left|\frac{1}{n}\sum_{i=1}^n \sigma_i f_j(X_i^{(j)})\right|.$$

Hence,
$$\mathbb{E}\left[\sup_f \frac{|1/n \sum_{i=1}^n \sigma_i f(X_i)|}{\lambda^{(2-\gamma)/2}\tau_{\text{tot}}(f)}\right]$$
$$\le \mathbb{E}\left[\sup_{f=\sum f_j} \sum_{j=1}^p \frac{|1/n \sum_{i=1}^n \sigma_i f_j(X_i^{(j)})|}{\lambda^{(2-\gamma)/2}\tau_{\text{tot}}(f)}\right]$$
$$= \mathbb{E}\left[\sup_{f=\sum f_j} \frac{1}{\tau_{\text{tot}}(f)} \sum_{j=1}^p \frac{|1/n \sum_{i=1}^n \sigma_i f_j(X_i^{(j)})|}{\lambda^{(2-\gamma)/2}\tau(f_j)}\tau(f_j)\right]$$
$$\le \mathbb{E}\left[\sup_{f=\sum f_j} \frac{1}{\tau_{\text{tot}}(f)} \max_j \sup_{\tilde f_j} \frac{|1/n \sum_{i=1}^n \sigma_i \tilde f_j(X_i^{(j)})|}{\lambda^{(2-\gamma)/2}\tau(\tilde f_j)} \sum_{j=1}^p \tau(f_j)\right]$$
$$= \mathbb{E}\left[\max_j \sup_{\tilde f_j} \frac{|1/n \sum_{i=1}^n \sigma_i \tilde f_j(X_i^{(j)})|}{\lambda^{(2-\gamma)/2}\tau(\tilde f_j)}\right]$$
$$\le \tilde C_s \frac{\delta_n}{\lambda} + c_{s,q}\lambda + \lambda^{\gamma/2}. \qquad \square$$

**B.6. Expectation of the weighted empirical process, indexed by the additive $f^2$'s.**

LEMMA 17. *Under Condition D,*
$$\mathbb{E}\left[\sup_f \frac{|\|f\|_n^2 - \|f\|^2|}{\tau_{\text{tot}}^2(f)}\right] \le 8c_{s,q}\lambda^{-\gamma/2}\mathbb{E}\left[\sup_f \frac{|1/n \sum_{i=1}^n \sigma_i f(X_i)|}{\tau_{\text{tot}}(f)}\right].$$

Proof. By a symmetrization argument (see, e.g., [31]),
$$\mathbb{E}\left[\sup_f \frac{|\|f\|_n^2 - \|f\|^2|}{\tau_{\text{tot}}^2(f)}\right] \le 2\mathbb{E}\left[\sup_f \frac{|1/n \sum_{i=1}^n \sigma_i f^2(X_i)|}{\tau_{\text{tot}}^2(f)}\right].$$

Because for all $j$,
$$\|f_j\|^\alpha I^{1-\alpha}(f_j) \le \lambda^{-\gamma/2}\tau(f_j),$$
we know from Corollary 7 that
$$|f_j|_\infty \le c_{s,q}\lambda^{-\gamma/2}\tau(f_j).$$

Hence,

$$|f|_\infty = \left|\sum_{j=1}^p f_j\right|_\infty \leq \sum_{j=1}^p |f_j|_\infty \leq c_{s,q}\lambda^{-\gamma/2}\sum_{j=1}^p \tau(f_j)$$
$$= c_{s,q}\lambda^{-\gamma/2}\tau(f).$$

Let $K = c_{s,q}\lambda^{-\gamma/2}$. Now, the function $x \mapsto x^2$ is Lipschitz on $[-K, K]$, with Lipschitz constant $2K$. Therefore, by the contraction inequality of Ledoux and Talagrand [18], we have

$$\mathbb{E}\left[\sup_f \frac{|1/n\sum_{i=1}^n \sigma_i f^2(X_i)|}{\tau_{\text{tot}}^2(f)}\right] \leq 4K\mathbb{E}\left[\sup_f \frac{|1/n\sum_{i=1}^n \sigma_i f(X_i)|}{\tau_{\text{tot}}(f)}\right]. \qquad \square$$

COROLLARY 8. *Using Lemma 16, we find under Condition D, and for* $\delta_n \leq \lambda \leq 1$, $\lambda \geq \sqrt{4(1+\log p)/n}$,

$$\mathbb{E}\left[\sup_f \frac{|\|f\|_n^2 - \|f\|^2|}{\tau_{\text{tot}}^2(f)}\right] \leq 8c_{s,q}\lambda^{1-\gamma}\left(\tilde{C}_s\frac{\delta_n}{\lambda} + c_{s,q}\lambda + \lambda^{\gamma/2}\right).$$

**B.7. Probability inequality for the weighted empirical process, indexed by the additive $f^2$'s.** We are now finally in the position to show that $\mathcal{S}_4$ has large probability.

THEOREM 6. *Let*

$$Z = \sup_f \frac{|\|f\|_n^2 - \|f\|^2|}{\tau^2(f)}.$$

*Assume Condition D, and* $\delta_n \leq \lambda \leq 1$, $\lambda \geq \sqrt{4(1+\log p)/n}$. *Then*

$$\mathbb{P}\left(Z \geq c_{s,q}\lambda^{1-\gamma}\left(2^7\tilde{C}_s\frac{\delta_n}{\lambda} + 32\lambda + 32\lambda^{\gamma/2} + \sqrt{2t}\right) + \frac{4c_{s,q}^2\lambda^{2(1-\gamma)}t}{3}\right)$$
$$\leq \exp(-n\lambda^{2-\gamma}t).$$

PROOF. We have

$$\frac{|f^2|_\infty}{\tau^2(f)} \leq c_{s,q}^2\lambda^{-\gamma}$$

and

$$\frac{\|f^2\|}{\tau^2(f)} \leq c_{s,q}\lambda^{-\gamma/2}\frac{\|f\|}{\tau(f)}$$

and

$$\|f\| \leq \sum_{j=1}^p \|f_j\| \leq \tau(f).$$

So we can apply the corollary of Bousquet's inequality with

$$K = c_{s,q}^2 \lambda^{-\gamma}$$

and

$$R = c_{s,q} \lambda^{-\gamma/2}.$$

We get that for all $t > 0$

$$\mathbb{P}\left(Z \geq 4\mathbb{E}[Z] + \frac{4c_{s,q}^2 t}{3n\lambda^{\gamma}} + c_{s,q}\sqrt{\frac{2t}{n\lambda^{-\gamma}}}\right) \leq \exp(-t).$$

Use the change of variable $t \mapsto n\lambda^{2-\gamma}t$, to reformulate this as: for all $t > 0$

$$\mathbb{P}\left(Z \geq 4\mathbb{E}[Z] + \frac{4c_{s,q}^2 \lambda^{2(1-\gamma)} t}{3} + c_{s,q}\lambda^{1-\gamma}\sqrt{2t}\right) \leq \exp(-n\lambda^{2-\gamma}t).$$

Now, insert

$$\mathbb{E}[Z] \leq 8c_{s,q}\lambda^{1-\gamma}\left(4\tilde{C}_s \frac{\delta_n}{\lambda} + c_{s,q}\lambda + \lambda^{\gamma/2}\right). \qquad \Box$$

REMARK 4. Recall that $\delta_n = \tilde{A}_s/\sqrt{n}$. Thus, taking $1 \geq \lambda \geq \tilde{A}_s/\sqrt{n}$ and $\lambda \geq \sqrt{4(1+\log p)/n}$, we see that for some constant $C_{s,q}$ depending only on $s$ and the lower bound for the marginal densities $\{q_j\}$, and for

$$C_0 = C_{s,q}(1 + \sqrt{2t} + \lambda^{1-\gamma}t),$$

we have

$$\mathbb{P}(\mathcal{S}_4) \geq 1 - \exp(-n\lambda^{2-\gamma}t).$$

**Acknowledgement.** We thank an Associate Editor and three referees for constructive comments.

L. MEIER
S. VAN DE GEER
P. BÜHLMANN
SEMINAR FÜR STATISTIK
ETH ZÜRICH
CH-8092 ZÜRICH
SWITZERLAND
E-MAIL: meier@stat.math.ethz.ch
        geer@stat.math.ethz.ch
        buhlmann@stat.math.ethz.ch